\documentclass{IEEEtran}
\usepackage{fancyhdr,xcolor}

\usepackage{amssymb}

\usepackage{amsthm}
\theoremstyle{plain}
\usepackage[compress]{cite}
\usepackage{graphics}
\usepackage{epstopdf}
\usepackage{epsfig}
\usepackage{lineno}
\usepackage{array}
\usepackage{amsmath,amssymb}
\usepackage{mathrsfs}
\usepackage{hyperref}
\usepackage{multirow}
\usepackage{bm}
\usepackage{bbm}
\usepackage{eqparbox}
\usepackage{algorithm}  
\usepackage{algorithmic}
\usepackage{graphicx}

\newtheorem{myRemark}{Remark}

\newtheorem{myDefinition}{Definition}
\usepackage{amsfonts}
\usepackage{enumerate}
\usepackage{indentfirst}
\usepackage{ragged2e}
\usepackage{url}
\usepackage{wrapfig}
\usepackage{soul}
\usepackage[labelformat=simple]{subcaption}

\usepackage{pifont}

\makeatletter
\newcommand*\rel@kern[1]{\kern#1\dimexpr\macc@kerna}
\newcommand*\widebar[1]{%
  \begingroup
  \def\mathaccent##1##2{%
    \rel@kern{0.8}%
    \overline{\rel@kern{-0.8}\macc@nucleus\rel@kern{0.2}}%
    \rel@kern{-0.2}%
  }%
  \macc@depth\@ne
  \let\math@bgroup\@empty \let\math@egroup\macc@set@skewchar
  \mathsurround\z@ \frozen@everymath{\mathgroup\macc@group\relax}%
  \macc@set@skewchar\relax
  \let\mathaccentV\macc@nested@a
  \macc@nested@a\relax111{#1}%
  \endgroup
}
\makeatother

\allowdisplaybreaks[4]

\begin{document}

\title{Positive-incentive Noise}

\author{Xuelong Li,\IEEEmembership{~Fellow,~IEEE}\\

\thanks{Xuelong Li is with \href{https://iopen.nwpu.edu.cn/en/Home.htm}{the School of Artificial Intelligence, OPtics and ElectroNics (iOPEN)}, 
Northwestern Polytechnical University, Xi'an 710072, Shaanxi, P. R. China.}

\thanks{
    This work is supported by The National Natural Science Foundation of China (No. 61871470).
}

\thanks{
    \copyright 2022 IEEE.  Personal use of this material is permitted.  Permission from IEEE must be obtained for all other uses, in any current or future media, including reprinting/republishing this material for advertising or promotional purposes, creating new collective works, for resale or redistribution to servers or lists, or reuse of any copyrighted component of this work in other works.
}

\thanks{E-mail: li@nwpu.edu.cn}

}

\pagestyle{fancy}


\renewcommand{\headrulewidth}{0pt}

\lhead{
    \footnotesize
    \makebox[\dimexpr\linewidth-2\fboxsep][l]{
        \textcolor{blue}{
        Xuelong Li, ``Positive-incentive Noise,'' \textit{IEEE Transactions on Neural Networks and Learning Systems}, DOI: 10.1109/TNNLS.2022.3224577. 
        \hfill
    }
    \thepage
  }
}


\maketitle

\thispagestyle{fancy}

\begin{abstract}
    Noise is conventionally viewed as a severe problem in diverse fields, 
    \textit{e.g.}, engineering, learning systems. 
    However, this paper aims to investigate whether the conventional proposition 
    always holds. It begins with the definition of task entropy, 
    which extends from the information entropy and measures the complexity of the task. 
    After introducing the task entropy, the noise can be classified into 
    two kinds, Positive-incentive noise (\textit{Pi-noise} or \textit{$\pi$-noise}) and pure noise, 
    according to whether the noise can reduce the complexity of the task. 
    Interestingly, as shown theoretically and empirically, even the simple 
    random noise can be the $\pi$-noise that simplifies the task. 
    $\pi$-noise offers new explanations for some models and provides a new principle for some fields, 
    such as multi-task learning, adversarial training, \textit{etc}. 
    Moreover, it reminds us to rethink the investigation of noises. 
\end{abstract}


\section{Introduction} \label{section_introduction}
Noise, which is conventionally regarded as a hurdle in pattern recognition
and machine learning, 
is ubiquitous due to a variety of reasons, \textit{e.g.}, 
human factors, instrumental error, and natural disturbances. 
Noise can be generated from different phases: 
(1) During the \textit{low-level data acquisition}, noises could come from instrumental 
errors; 
(2) at the \textit{data level}, noises may be caused by the differences of data storage and representation; 
(3) at the \textit{feature level}, noises are usually generated by the imprecise modelings; 
(4) there may exist \textit{instance-level} noises as well, \textit{i.e.}, 
irrelevant data points. 
There is a potential assumption in existing works: 
\textit{the noise \textbf{always} causes a negative impact to 
the current task.}  
Therefore, how to design a model insensitive to noise is an important 
topic in various fields of pattern recognition. 
For example, in computer vision, plenty of filters are designed to 
alleviate the impact of noise, \textit{e.g.}, 
Gaussian filter, uniform filter. 
In the past decade with the rapid growth of machine learning, 
the robust model is an extremely studied topic, 
\textit{e.g.}, 
noise-insensitive clustering \cite{RobustClustering}, 
robust feature selection \cite{FS}, 
multi-view learning \cite{MJP}, 
noisy matrix completion \cite{NCARL}, 
adversarial training \cite{AT}. 

Nevertheless, does the above assumption always holds?  
Or formally, the crucial question that this paper intends to answer 
is: 
\textbf{\textit{is noise always harmful?}}

The question originates from some inspiring instances of noise. 
The first one is the \textit{traffic noises} from the cars for acoustics tasks. 
In most scenes, the car noises should be regarded as useless information 
or disturbance due to the unsatisfied reception when collecting data. 
However, if the acoustics task is relevant to time, the car noise 
may offer extra information about time and \textit{enhance} performance. 
Generally speaking, the intensities of car noises in the morning rush hour and midnight 
are clearly different, which can provide coarse information about time.  
Another inspiring instance is the \textit{gum example}. 
For a clean wall, either gum or nail is a kind of noise for the wall. 
It means that the gum stuck on the wall and embedded nail are both unexpected. 
However, a piece of gum may help to remove the embedded nail with the help 
of its adhesive ability. 
Although there will be a hole, the gum, a kind of noise, is used to \textit{rectify} another noise. 
Or similarly, the gum may help to remove a broken key stuck in a lock, 
while both the gum and the broken key are noise for the lock. 

Inspired by the above instances, a question comes: \textit{do noises really mislead 
the target task in all cases?} 
A topic related to the question is stochastic resonance \cite{SR}, 
which employs random noises to enhance the detection of weak signals. 
However, SR fails to completely answer the above question since it only 
focuses on the scenes about weak signal detection. 

The crucial factor causing doubt about noises is the loose definition of noise. 
To rigorously answer the question, the complexity (or equivalently difficulty, uncertainty) 
of the given \textit{task} plays an important role. 
With the definition of task entropy, the conventionally defined noise can be classified into 
2 categories. One is the noise decreasing the complexity of the task, namely 
Positive-incentive noise (\textit{Pi-noise} or \textit{$\pi$-noise}). 
Another is the useless noise for the task, namely pure noise. 
With the proper definition of task entropy, stochastic resonance is 
a specific case of $\pi$-noise. 
Several subfields of pattern recognition (\textit{e.g.}, multi-task learning, 
adversarial training) are also connected with $\pi$-noise. 
It should be emphasized that \textit{superfluous $\pi$-noises also result in negative impact}, 
which is the reason why it is still named ``noise''. 
For instance, car noises will also disturb time-related acoustics recognition 
if the noises are too strong. 

In the following part, Section \ref{section_preliminary} introduces the 
mathematical notations appearing in this paper. 
Section \ref{section_pi_noise} elaborates on the theoretical motivation and 
definition of $\pi$-noise. In Section \ref{section_experiments}, 
two applicable topics of $\pi$-noise are discussed and sufficient experiments 
also verify the existence and effectiveness of $\pi$-noise. 
The experimental results provide a counterintuitive conclusion: 
Even a simple random noise may simplify the task with the proper setting.

\section{Preliminary} \label{section_preliminary}
In this paper, matrices and vectors are denoted by uppercase and 
lowercase letters in boldface, respectively. 
The information entropy \cite{Entropy} of a random variable $x$ is denoted by 
\begin{equation}
    H(x) = 
    \begin{cases}
    - \int p(x) \log p(x) dx & \textrm{if } x \textrm{ is continuous} \\
       - \sum_{x} p(x) \log p(x) & \textrm{if } x \textrm{ is discrete}
    \end{cases}
    .
\end{equation}
And the mutual information of two discrete random variables is computed by \cite{InformationTheoryBook} 
\begin{equation}
    \begin{split}
        \textrm{MI}(x, y) & = \sum_{x, y} p(x, y) \log \frac{p(x, y)}{p(x) p(y)} \\ 
        & = H(x) - H(x | y), 
    \end{split}
\end{equation}
where the conditional entropy is defined as 
\begin{equation}
    H(x | y) = - \sum p(x, y) \log p(x | y) . 
\end{equation}
The above definition can be easily extended to continuous variables 
by replacing the sum operator with the integral symbol. 
$\delta(x)$ is the Dirac delta function. 
$\textrm{sgn}(x)$ returns $x/|x|$ if $x \neq 0$ and 0 otherwise. 
The noise is denoted by $\bm \epsilon$ if without any specific statement.

\section{Positive-incentive Noise} \label{section_pi_noise}
In this section, the motivation and the formal definition of $\pi$-noise 
are introduced first. 
Then, the relations between some existing fields and $\pi$-noise are 
elaborated. 

\begin{figure}[t]
    \centering
    \subcaptionbox{Sophisticated Image}{
        \includegraphics[width=0.46\linewidth]{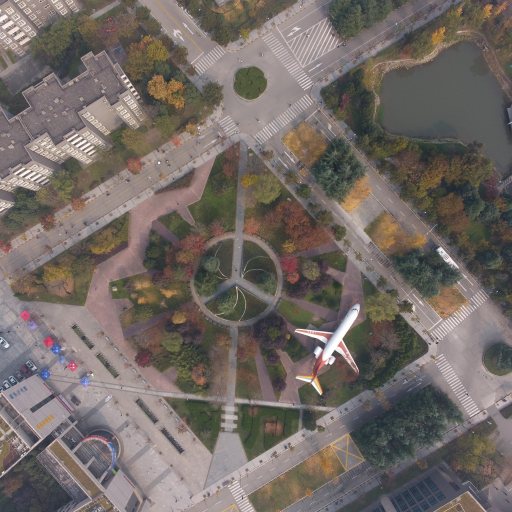}
    }
    \subcaptionbox{Simple Image \label{figure_coil20}}{
        \includegraphics[width=0.46\linewidth]{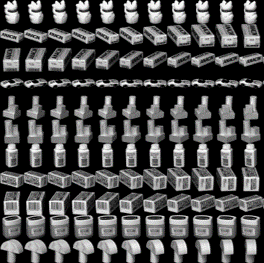}
    }
    \caption{Comparison between sophisticated image and simple image. 
    The left one is an aerial image and the right one is COIL20 \cite{COIL20}.
    For single-label classification, the left one can be labeled as ``plane'', ``building'', 
    or ``tree''. The uncertainty increases with growth of label space $\mathcal{Y}$ 
    due to the abundant information in the image. }
    \label{figure_two_pictures}
\end{figure}

\subsection{Motivation from Information Theory}
The behind philosophy of $\pi$-noise is that the same noise may play different roles in 
diverse tasks. The inspiring example is also about the car noise. 
For most acoustics recognition tasks, car noise is the unexpected 
additive signal caused by unsatisfied reception. 
For time-relevant tasks, however, the car noise provides the extra beneficial 
information. 

Accordingly, it implies that the rigorous discussion of noise should be based 
on \textit{tasks}. 
Before discussing the relationship between task and noise, 
how to mathematically measure a task $\mathcal{T}$ is the first crucial question. 
With the help of information theory, the entropy of $\mathcal{T}$ can be 
defined to indicate the complexity of $\mathcal{T}$. 
Formally speaking, 
the smaller $H(\mathcal{T})$ means the easier task. 
Clearly, how to compute $H(\mathcal{T})$ is the key problem. 
In the following part,  the rationality of $H(\mathcal{T})$ and 
how to compute it are shown with the help of a general classification task.

If the entropy of task $\mathcal{T}$ can be formulated, 
it is natural to define the mutual information of task $\mathcal{T}$ and 
noise $\bm \epsilon$, 
\begin{equation}
    {\rm MI} (\mathcal{T}, \bm \epsilon) = H(\mathcal{T}) - H(\mathcal{T} | \bm \epsilon) . 
\end{equation}
In the context of the conventional discussion of noise, 
the strict definition of unexpected and harmful noise should satisfy 
${\rm MI} (\mathcal{T}, \bm \epsilon) = 0$. 
However, as theoretically and empirically shown in this paper, 
even the simple random noise (\textit{e.g.}, Gaussian noise) may lead to positive mutual information. 
That is an interesting phenomenon since it implies that finding completely 
unrelated random noise may be also difficult. 
Formally, the definition of $\pi$-noise is given as follows: 
\begin{myDefinition}
Formally, define the noise $\bm \epsilon$ satisfying the following condition, 
\begin{equation}
    {\rm MI} (\mathcal{T}, \bm \epsilon) > 0,
\end{equation}
as the \textit{$\pi$-noise}. 
The above inequality is also equivalent to 
\begin{equation}
    H(\mathcal{T}) > H(\mathcal{T} | \bm \epsilon) , 
\end{equation}
which indicates that $\bm \epsilon$ simplifies the original task. 
On the contrary, the noise satisfying ${\rm MI}(\mathcal{T}, \bm \epsilon) = 0$ is 
named the \textit{negative noise} or \textit{pure noise}. 
\end{myDefinition}

Furthermore, more strict $\pi$-noise can be defined by introducing a threshold:
\begin{myDefinition}
	The noise $\bm \epsilon$ is named as $\alpha$-strong $\pi$-noise if it satisfies 
	\begin{equation}
		{\rm MI}(\mathcal{T}, \bm \epsilon) > \alpha . 
	\end{equation}
	Similarly, the noise satisfying ${\rm MI}(\mathcal{T}, \bm \epsilon) \leq \alpha$ 
	is named as $\alpha$-strong negative/pure noise.
\end{myDefinition}

It should be emphasized that $\pi$-noise can be viewed as 
a kind of information gain brought by $\bm \epsilon$. 
One may argue why not to define $\pi$-noise via the information gain, 
which is widely used in machine learning. 
The mutual information is preferable due to that it directly shows the essence of $\pi$-noise. 
In other words, the random noise component contains useful information for $\mathcal{T}$. 
Various measurements of the information gain could be used 
to estimate ${\rm MI}(\mathcal{T}, \bm \epsilon)$ and help to 
distinguish the $\pi$-noise.

\begin{myRemark}[Moderate $\pi$-Noise Assumption]
    The existence of $\pi$-noise does not indicate that there exists a random 
    variable so that $\mathcal{T}$ can be persistently enhanced with the 
    increase of the $\pi$-noise. 
    Even for the $\pi$-noise, the conventional consensus of noise does not change: 
    superfluous $\pi$-noise will cause degeneration. 
    In other words, $\bm \epsilon = 0$ holds in most cases. 
    This is the reason why $\pi$-noise is still named ``noise''. 
\end{myRemark}

As shown in the following subsections, some existing relevant topics can be 
viewed as special cases of the $\pi$-noise framework.

\subsection{Explanation of Single-Label Classification} \label{section_pi_noise_classification}
For a fundamental single-label classification problem, the dataset $(\bm X, \bm Y)$ 
can be regarded as samplings from $\mathcal{D}_{\mathcal{X}, \mathcal{Y}}$ \cite{UnderstandingML}
where $\mathcal{D}_{\mathcal{X}, \mathcal{Y}}$ is the underlying 
joint distribution of data points and labels from feasible space  
$\mathcal{X}$ and $\mathcal{Y}$, 
\textit{i.e.}, $(\bm X, \bm Y) \sim \mathcal{D}_{\mathcal{X}, \mathcal{Y}}$. 
Therefore, 
given a set of data points $\bm X$, the label set can be regarded 
as sampling from $\bm Y \sim \mathcal{D}_{\mathcal{Y} | \mathcal{X}}$ and 
the ``\textit{complexity}'' (equivalently \textit{difficulty} or \textit{uncertainty}) of $\mathcal{T}$ on $\bm X$ is formulated as 
\begin{equation}
    H(\mathcal{T}; \bm X) = - \sum_{\bm Y \in \mathcal{Y}} p(\bm Y | \bm X) \log p(\bm Y | \bm X) . 
\end{equation}
To better understand $p(\bm Y | \bm X)$, 
Fig. \ref{figure_two_pictures} shows two image datasets for the classification tasks. 
The left one is an aerial image where the label space is 
subjected to $\mathcal{Y} = \{\textrm{plane, building, tree}, \ldots\}$. 
It may be therefore tagged as ``plane'', ``building'', or ``tree''. 
which is also the intention of label smoothing \cite{LabelSmoothing}.
The uncertain label increases the complexity of $\mathcal{T}$. 
On the contrary, classification on the pure images of objects without complicated background, 
\textit{e.g.}, COIL20 \cite{COIL20} shown in Fig. \ref{figure_coil20}, is 
is more simple task. 
If $H(\mathcal{T}; \bm X) = 0$ holds, it indicates that 
there exists a $\bm Y_*$ so that 
\begin{equation}
    p(\bm Y | \bm X) = 
    \begin{cases}
        1 & \bm Y = \bm Y_* \\
        0 & \textrm{else}
    \end{cases}
    . 
\end{equation}
In this case, the task is apparently the simplest since all semantic ambiguity 
does not exist. 
Furthermore, the expected entropy of task $\mathcal{T}$ (\textit{i.e.}, 
independent of some specific dataset $\bm X$) can be defined as 
\begin{equation}
    H(\mathcal{T}) = \mathbb{E}_{\bm X \sim \mathcal{D}_{\mathcal{X}}} H(\mathcal{T}; \bm X) . 
\end{equation}
Note that the expected task entropy $H(\mathcal{T})$ is actually the conditional entropy 
$H(\mathcal{T} | \bm X)$. To keep simplicity and generality, 
$H(\mathcal{T})$ is instead used. 
In the following part, both the specific task entropy and 
the expected entropy are denoted by $H(\mathcal{T})$ if unnecessary.
\textit{Another interesting corollary is that the task entropy can 
measure the quantity of information under the context of the classification 
task to some extent.} 

\subsection{Explanation of Stochastic Resonance}
Stochastic resonance (\textit{SR}) \cite{SR}, 
which is known as a kind of noise benefit, 
is firstly discussed to provide a specific instance. 
For signal $y_t = f(t) \in \Theta$, the assumption of SR is the weak stimuli and 
$f(t) < \theta$ holds in most cases where $\theta$ represents the minimum 
threshold that sensors can detect. 
The goal of signal detection is to detect the weak signal as much as possible. 
The unseen stimuli imply that any value in the feasible domain may be possible 
which leads to strong randomness. 
Formally speaking, 
define $p(y_t)$ as
\begin{equation}
    p(y_t = s | t = T) = 
    \begin{cases}
        \frac{1}{\theta - \theta_0} & T \notin \mathcal{S}_f \\
        \delta (s - f_o(T))  & T \in \mathcal{S}_f
    \end{cases}
    ,
\end{equation}
where $\theta_0 = \inf_{y_t \in \Theta} y_t$, $\mathcal{S}_f = \{t | y_t \geq \theta\}$, 
and $f_o(t)$ is the observed quantity. 
Accordingly, the task entropy can be formulated as 
\begin{equation}
    H(\mathcal{T}_{\rm \textit{SR}}) = H(y_t | t) = \iint - p(y_t, t) \log p(y_t | t) dy_t dt . 
\end{equation}
Clearly, if $\mathcal{S}_f = \emptyset$, $p(y_t)$ is a uniform distribution 
and $H(\mathcal{T}_{\rm \textit{SR}})$ achieves the maximum of the entropy. 
Provided that $\epsilon_t \sim \mathcal{N}(0, \sigma^2)$, 
the joint probability and conditional probability can be formulated as 
\begin{equation}
    p(y_t = s | \epsilon_t = \epsilon_0, t = T) = 
    \begin{cases}
        \frac{1}{\theta - \theta_0} & T \notin \mathcal{S}_{f + \epsilon} \\
        \delta (s - f_o(t) + \epsilon_0)  & T \in \mathcal{S}_{f + \epsilon}
    \end{cases}
    ,
\end{equation}
and 
\begin{equation}
	\begin{split}
    	p(y_t, \epsilon_t, t) & = p(y_t | \epsilon_t, t) \cdot p(\epsilon_t | t) \cdot p(t)\\
		& = p(y_t | \epsilon_t, t) \cdot \mathcal{N}(0, \sigma^2) \cdot p(t),
	\end{split}
\end{equation}
where $\mathcal{S}_{f + \epsilon} = \{t | y_t + \epsilon > \theta\}$. 
Accordingly, the conditional entropy is formulated as 
\begin{equation}
	\begin{split}
    H(\mathcal{T}_{\rm \textit{SR}} | \bm \epsilon) & = \iiint -p(y_t, \epsilon_t, t) \log p(y_t | t, \epsilon_t) dy_t d\epsilon_t dt \\
	& = \iiint -p(y_t | \epsilon_t, t) \cdot \mathcal{N}(0, \sigma^2) \cdot p(t) dy_t d\epsilon_t dt . 
	\end{split}
\end{equation}
Consider an extreme example when $\mathcal{S}_f = \emptyset$ and the following inequality, 
\begin{equation}
    H(\mathcal{T}_{\rm \textit{SR}}) > H(\mathcal{T}_{\rm \textit{SR}} | \bm \epsilon), 
\end{equation}
easily holds if $\sigma$ is appropriate. 
On the contrary, when $\widebar{\mathcal{S}}_f = \emptyset$ (where $\widebar{\mathcal{S}}_f$ represents the complementary set of $\mathcal{S}_f$), 
it is also easy to 
obtain 
\begin{equation}
    H(\mathcal{T}_{\rm \textit{SR}}) = H(\mathcal{T}_{\rm \textit{SR}} | \bm \epsilon) . 
\end{equation}
The analysis of SR shows that the random noise $\bm \epsilon$ may be 
$\pi$-noise on some data but be pure noise in other cases. 
It may implies that there exist no noise being $\pi$-noise or pure noise on 
every dataset for task $\mathcal{T}$. 

\subsection{Explanation of Multi-Task Learning}
Multi-task learning \cite{LowRankMultiTask} could be regarded as a special case of $\pi$-noise.
If $\bm \epsilon$ represents one or some tasks, \textit{i.e.}, 
$\bm \epsilon = (\mathcal{T}_1, \mathcal{T}_2, \ldots, \mathcal{T}_k)$. 
Suppose that all tasks related to $\mathcal{T}$ are denoted by $\mathcal{G}$ 
and other irrelevant tasks are denoted by $\widebar{\mathcal{G}}$. 
The low-rank multi-task models \cite{LowRankMultiTask} intend to employ $\{\mathcal{T}\} \cup \mathcal{G}$ 
and eliminate $\widebar{\mathcal{G}}$, 
since $\mathcal{G}$ is the $\pi$-noise and $\widebar{\mathcal{G}}$ is pure noise. 
In other words, $H(\mathcal{T} | \mathcal{G}) < H(\mathcal{T})$ 
is the reason \textit{why the multi-task learning outperforms the original task.}

\begin{figure}
	\centering

    \subcaptionbox{\label{pepper noise 0}Original}{
        \includegraphics[width=0.21\linewidth]{./figure/NWPU/p2/x_2_small.png}
    }
    \subcaptionbox{\label{pepper noise 0.1}Degree=0.1}{
        \includegraphics[width=0.21\linewidth]{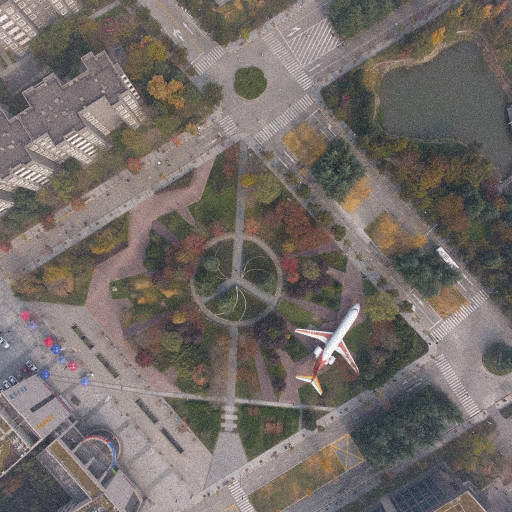}
    }
    \subcaptionbox{\label{pepper noise 0.3}Degree=0.3}{
        \includegraphics[width=0.21\linewidth]{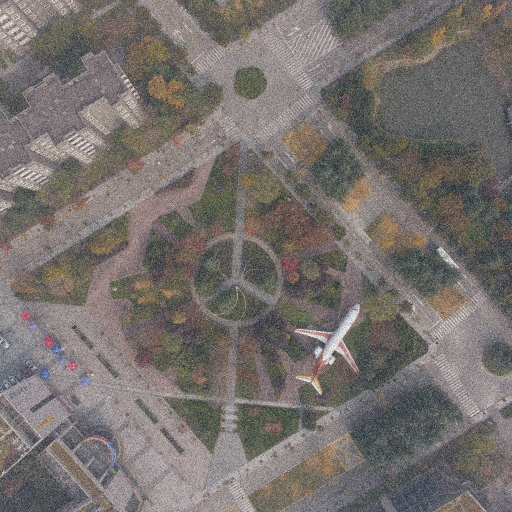}
    }
    \subcaptionbox{\label{pepper noise 0.5}Degree=0.5}{
        \includegraphics[width=0.21\linewidth]{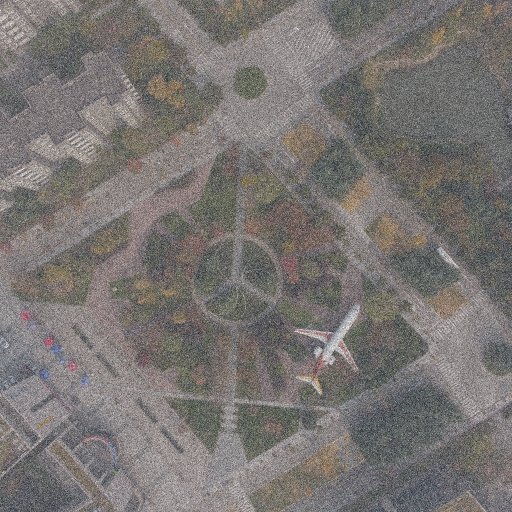}
    }
	
    \subcaptionbox{\label{Gaussian Original}Original}{
        \includegraphics[width=0.21\linewidth]{./figure/NWPU/p2/x_2_small.png}
		
    }
    \subcaptionbox{\label{gaussian mean_0_0_sigma_0_5}\scriptsize{$\mu$=$0.0$ $\sigma$=$0.5$}}{
        \includegraphics[width=0.21\linewidth]{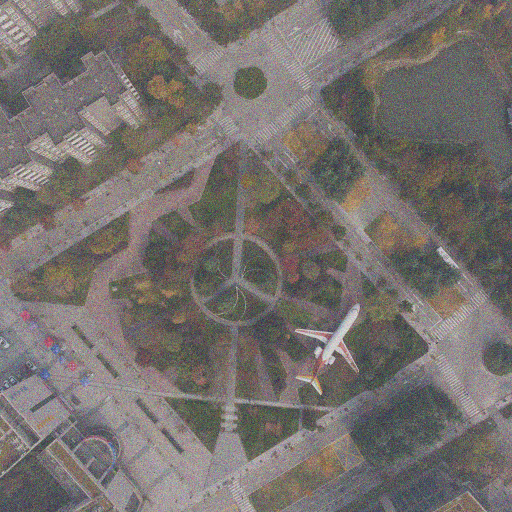}
    }
    \subcaptionbox{\label{gaussian mean_0_5_sigma_0_0}\scriptsize{$\mu$=$0.5$ $\sigma$=$0.0$}}{
        \includegraphics[width=0.21\linewidth]{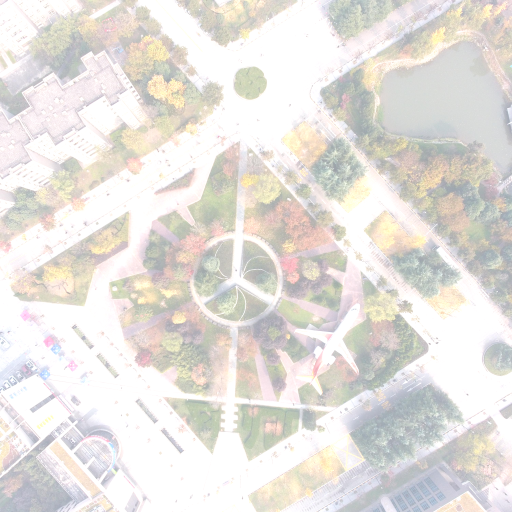}
    }
    \subcaptionbox{\label{gaussian mean_0_5_sigma_0_5}\scriptsize{$\mu$=$0.5$ $\sigma$=$0.5$}}{
        \includegraphics[width=0.21\linewidth]{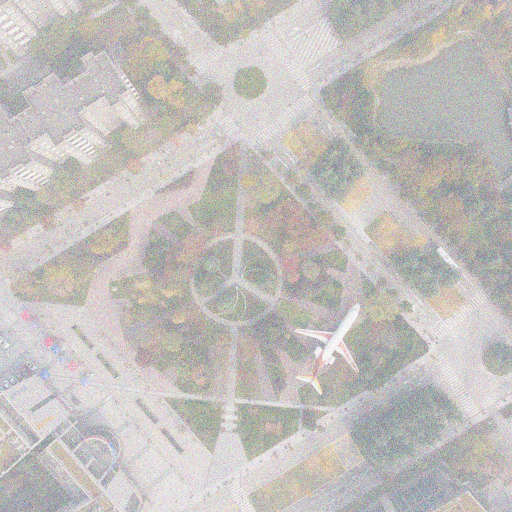}
    }

    \subcaptionbox{\label{Uniform Original}Original}{
        \includegraphics[width=0.21\linewidth]{./figure/NWPU/p2/x_2_small.png}
    }
    \subcaptionbox{\label{uniform low_0_high_0_3}\scriptsize{$a$=$0.0$ $b$=$0.3$}}{
        \includegraphics[width=0.21\linewidth]{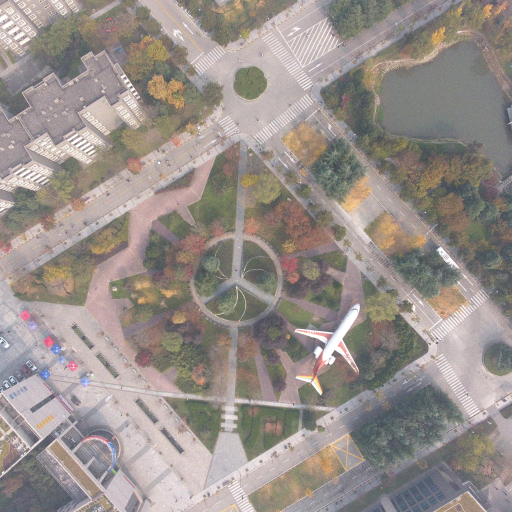}
    }
    \subcaptionbox{\label{uniform low_0_high_0_5}\scriptsize{$a$=$0.0$ $b$=$0.5$}}{
        \includegraphics[width=0.21\linewidth]{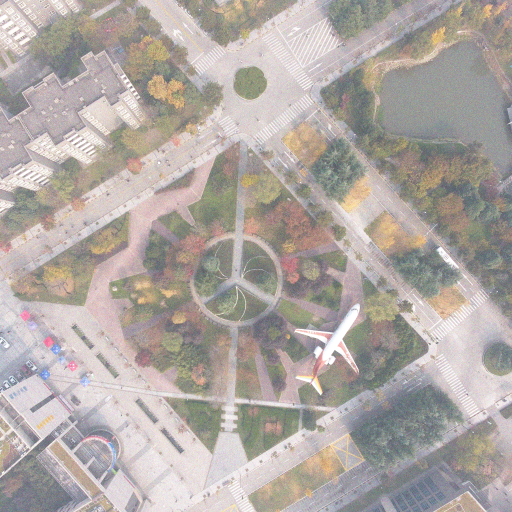}
    }
    \subcaptionbox{\label{uniform low_0_high_1}\scriptsize{$a$=$0.0$ $b$=$1.0$}}{
        \includegraphics[width=0.21\linewidth]{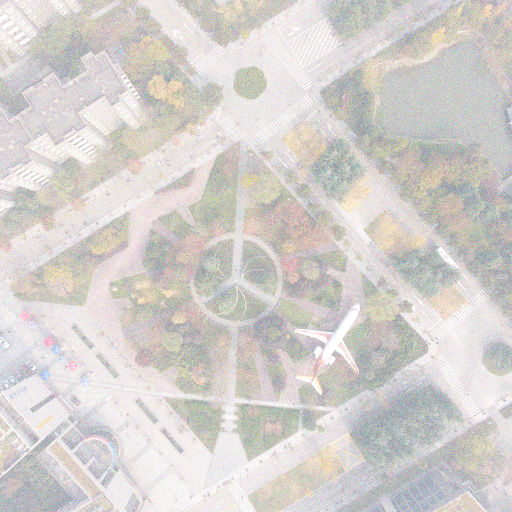}
    }

	\caption{The visualization of images with three different kinds of noise. 
		The first row shows the images with four degrees of multiplicative noise. 
		The second row suggests three Gaussian noisy images. 
		The bottom row shows the image with three kinds of uniform noise. 
		The moderate noise blurs the background so that $H(\mathcal{T} | \bm \epsilon)$ 
		is smaller than $H(\mathcal{T})$. }
	\label{Noise Image}
\end{figure}

\begin{figure*}[t]
    \centering
    \subcaptionbox{Multiplicative Noise \label{Multiplicative Noise}}{
        \centering
        \includegraphics[width=0.31\linewidth]{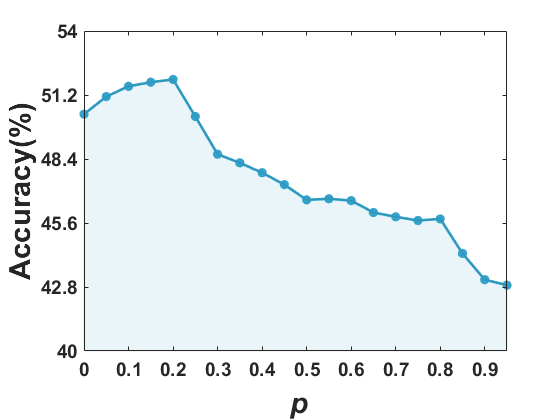}
    }
    \subcaptionbox{Gaussian Noise \label{Gaussian Noise}}{
        \centering
        \includegraphics[width=0.31\linewidth]{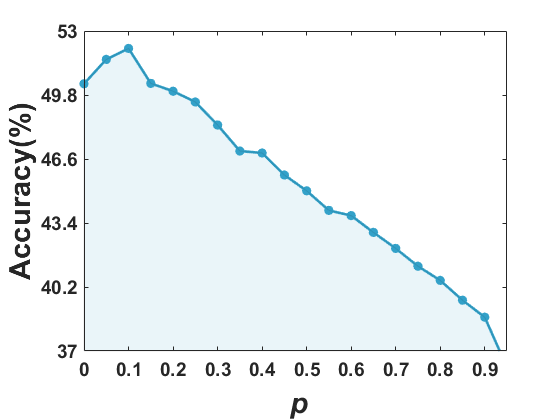}
    }
    \subcaptionbox{Uniform Noise \label{Uniform Noise}}{
        \centering
        \includegraphics[width=0.31\linewidth]{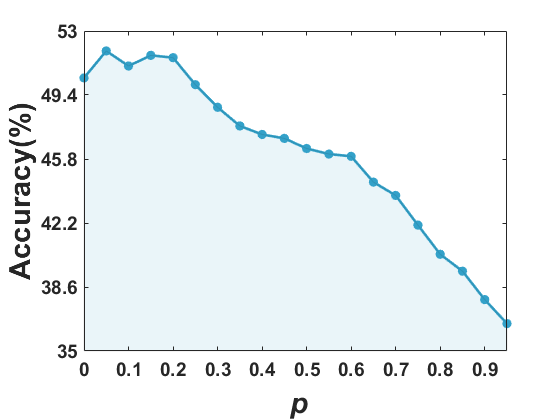}
    }
    \caption{The classification accuracy on noisy image dataset with a different noisy ratio $p$. X-axis is a noisy ratio $p$ and Y-axis is a classification accuracy. Fig. \subref{Multiplicative Noise}-\subref{Uniform Noise} show the results with multiplicative noise, Gaussian noise, and uniform noise, respectively.}
    \label{Classification Accuracy}
\end{figure*}

\subsection{Relationship Between $\pi$-Noise and Adversarial Training}
$\pi$-noise also offers a new perspective for the adversarial training, 
which seems relevant to $\pi$-noise framework. 
The adversarial training \cite{AT} is usually formulated as 
\begin{equation}
    \min_{\theta} \sum_{\bm x \in \bm X} \max _{\|\bm \epsilon\| \leq C} \ell(f_{\bm \theta}(\bm x + \bm \epsilon), \bm y) , 
\end{equation}
where $\ell(\cdot)$ represents some loss function, $\bm \theta$ is the learning parameters of the model and 
$C$ is a constant. 
The goal of adversarial training is to \textit{enhance the robustness of the model} 
$f_{\bm \theta}$ via introducing the adversarial perturbation $\bm \epsilon$. 
In other words, the underlying assumption is: $f_{\bm \theta}$ achieves 
satisfying performance on $\bm X$ but obtains unexpected generalization performance. 

In the $\pi$-noise framework, $\bm \epsilon$ is used to \textit{reduce the complexity of the task}, 
instead of aiming at any specific models. 
More precisely, the purpose of introducing $\pi$-noise is to decrease the difficulty 
of training any models. 
Large $H(\mathcal{T})$ usually implies that a model probably learns imprecise 
semantic information, which may provide new perspectives to understand why some models are not stable 
on complicated datasets. 
For instance, the classification models may over-evaluate those points with uncertain labels. 



\section{Applications of $\pi$-Noise} \label{section_experiments}
After theoretically discussing the definition of $\pi$-noise (and pure noise),  
two possible applications of $\pi$-noise are provided in this section. 
Some experiments are also conducted to show the universal existence of 
$\pi$-noise.

\subsection{Enhanced $\pi$-Noise}
The first application is to use $\pi$-noise to enhance the performance 
which is direct from the definition and corresponds to 
multi-task learning. 
Rigorously speaking, the enhancement of performance is based on 
\textit{decreasing the complexity of tasks via $\pi$-noise}. 
This part of the experiment is also a direct answer to the question proposed 
in the title: 
\textit{Even for the simple random noise, the impact is not always negative}.

\begin{table}[t]
    \centering
    \renewcommand\arraystretch{1.2}
    \caption{Datasets}
    \label{table_datasets1}
    \begin{tabular}{lccc}
        \hline
        
        \hline
        \textbf{Datasets} & \textbf{\# Samples} & \textbf{\# Features} & \textbf{\# Classes} \\ 
        \hline
        Cars               & 392                    & 8                 & 3                \\
        Balance           & 624                    & 4                & 3               \\
        Australian        & 690                    & 14                 & 2               \\
        Breast           & 699                    & 10                & 2                \\
        Diabets       & 768                    & 8                & 2               \\
        \hline
        
        \hline
    \end{tabular}
\end{table}

\subsubsection{Datasets Setting}

The experiments of enhanced $\pi$-noise are conducted on the image classification 
task. 
The real image dataset STL-10 \cite{STL10} is chosen as the benchmark dataset. 
This dataset has $10$ class samples. 
Each class has $500$ training images and $800$ testing images. 
Suppose that the original image is noiseless and three categories of 
noise (including multiplicative noise, Gaussian noise, and uniform noise) 
are added to the data before training. For the dimension noise, five UCI benchmark are selected and the details are listed in Table \ref{table_datasets1}. $50\%$ of the sampled points from each class are employed as the training data and the rest are acted as the test data.
Meanwhile, LeNet \cite{LeNet} is chosen as the baseline method to extract the deep feature and output the predicted classification. 
This network is trained by the stochastic gradient descent to minimize the cross-entropy loss. 
Besides, the batch size is $200$ and the epoch is $50$. 
The learning rate is $0.01$. Furthermore, SVM \cite{SVM}, Lasso \cite{Lasso}, and DLSR \cite{DLSR} are employed as the classifier to evaluate the performance. 
Among them, the regularization parameter of Lasso and DLSR is set to $0.01$ and $1$, respectively. 
The classification accuracy (ACC) metric is employed to evaluate the performance of the network.

\subsubsection{Details of Generated Noise}

To be more persuasive, four categories of noises are applied to the original training set. 
In particular, the first three kinds of noise (multiplicative noise, Gaussian noise, and uniform noise) 
are generated according to the different proportion $p=\frac{N_{\bm \epsilon}}{N}$, where $N_{\bm \epsilon}$ is the number of noisy training samples and $N$ is the total number of training samples. 
The detailed settings of the noises are listed as follows:

\begin{table}[t]
    \renewcommand\arraystretch{1.3}
    \centering
    \caption{Extension Results on Dimension Noise}
    \label{demension_results}
    \scalebox{0.80}{
        \begin{tabular}{lccccccccc}
            \hline
            
            \hline
            \multicolumn{1}{l}{}    & \multicolumn{3}{c}{\textbf{SVM}}  & \multicolumn{3}{c}{\textbf{Lasso}} & \multicolumn{3}{c}{\textbf{DLSR}}    \\ \cline{2-4}  \cline{5-7} \cline{8-10} 
            \multicolumn{1}{l}{\multirow{-2}{*}{\textbf{}}} 
            & $m$ & \emph{ACC}  & \emph{Pi-ACC}  & $m$ & \emph{ACC} & \emph{Pi-ACC}  & $m$ & \emph{ACC} & \emph{Pi-ACC} \\ \hline
            \textbf{Cars}            & 6 & 64.76 & \textbf{69.52} & 10 & 75.90 & \textbf{81.03} & 16 & 78.46 & \textbf{82.05} \\
            \textbf{Balance} & 10 & 91.35 & \textbf{91.67} & 18 & 89.10 & 89.10 & 14 & 87.18 & \textbf{89.42}  \\
            \textbf{Australian}  & 20 & 85.76 & \textbf{87.79} & 20 & 87.21 & \textbf{88.66} & 8 & 86.92 & \textbf{88.37}  \\
            \textbf{Breast} & 12 & 95.70 & \textbf{96.56} & 12 & 94.84 & \textbf{97.99} & 12 & 95.13 & \textbf{97.13}  \\ 
            \textbf{Diabetes} & 6 & 74.74 & \textbf{76.82} & 14 & 75.52 & \textbf{76.30} & 4 & 75.52 & \textbf{76.82}  \\
            \hline
            
            \hline
        \end{tabular}
    }
\end{table}

\begin{figure*}[t]
    \centering
    \subcaptionbox{Toy ACC: 100.00\% \label{Toy SVM}}{
        \centering
        \includegraphics[scale=0.28]{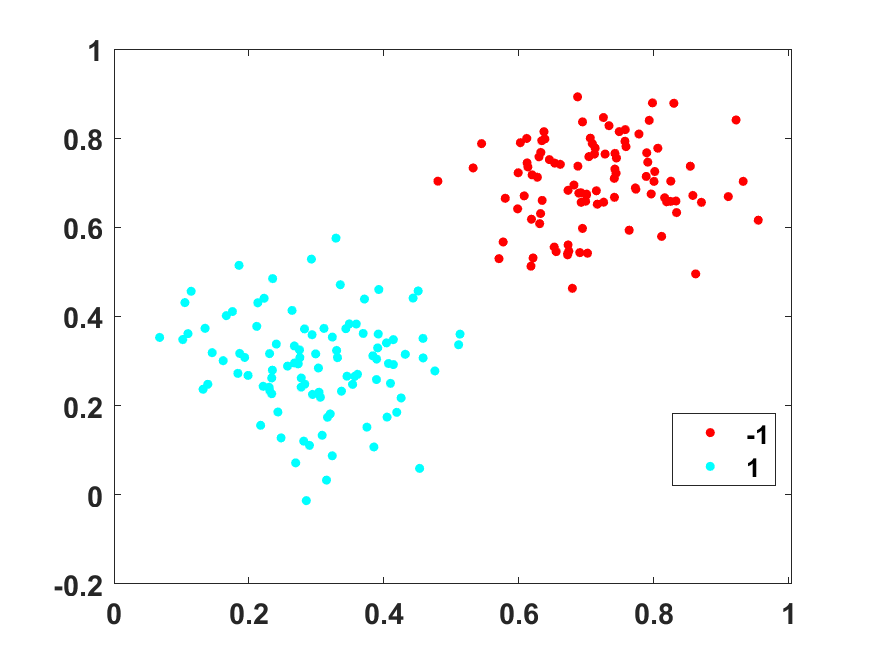}
    }
    \subcaptionbox{Noise Suppression: 91.82\% \label{Toy Noise SVM}}{
        \centering
        \includegraphics[scale=0.28]{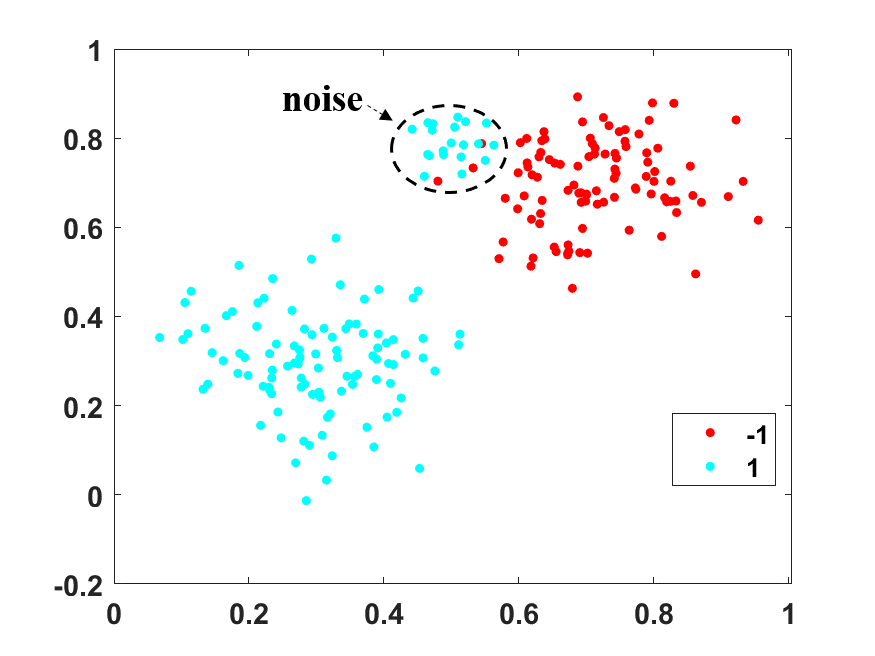}
    }
    \subcaptionbox{Rectified by $\pi$-noise: 92.80\% \label{Toy Pi-noise SVM}}{
        \centering
        \includegraphics[scale=0.28]{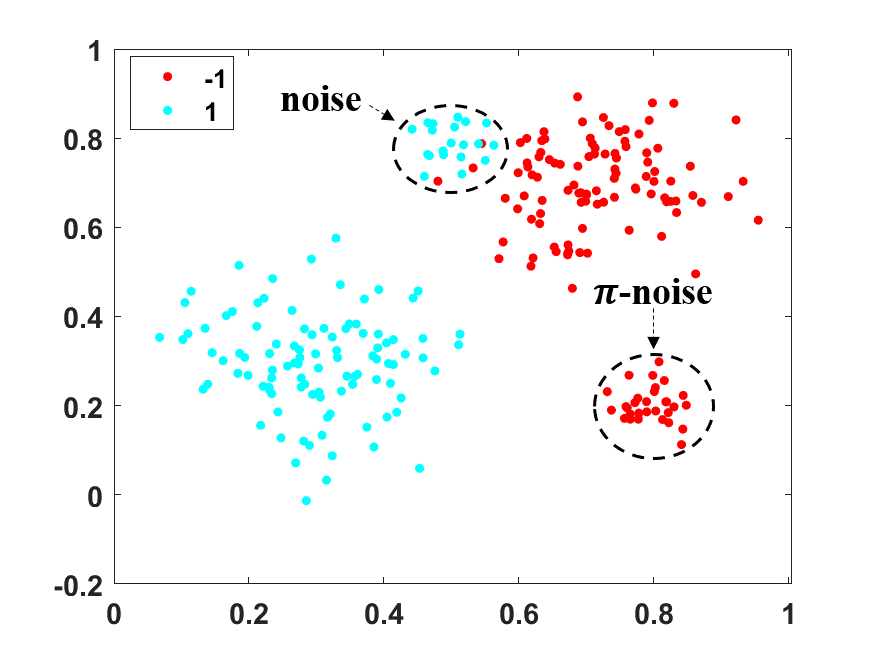}
    }
    \subcaptionbox{Excessive $\pi$-noise: 93.46\% \label{Toy over Pi-noise SVM}}{
        \centering
        \includegraphics[scale=0.28]{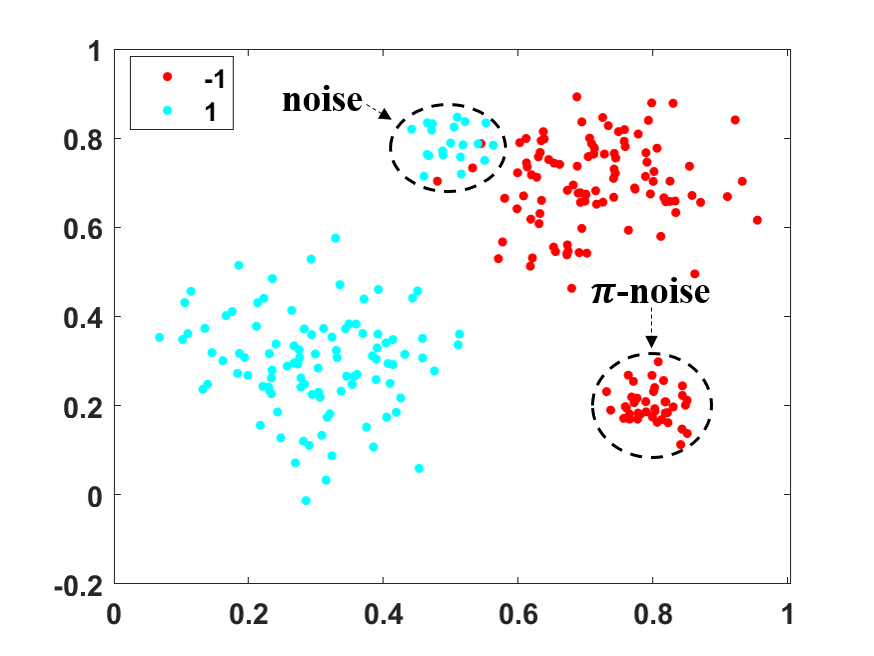}
    }
    
    
    \subcaptionbox{Iris ACC: 96.67\% \label{Iris SVM}}{
        \centering
        \includegraphics[scale=0.28]{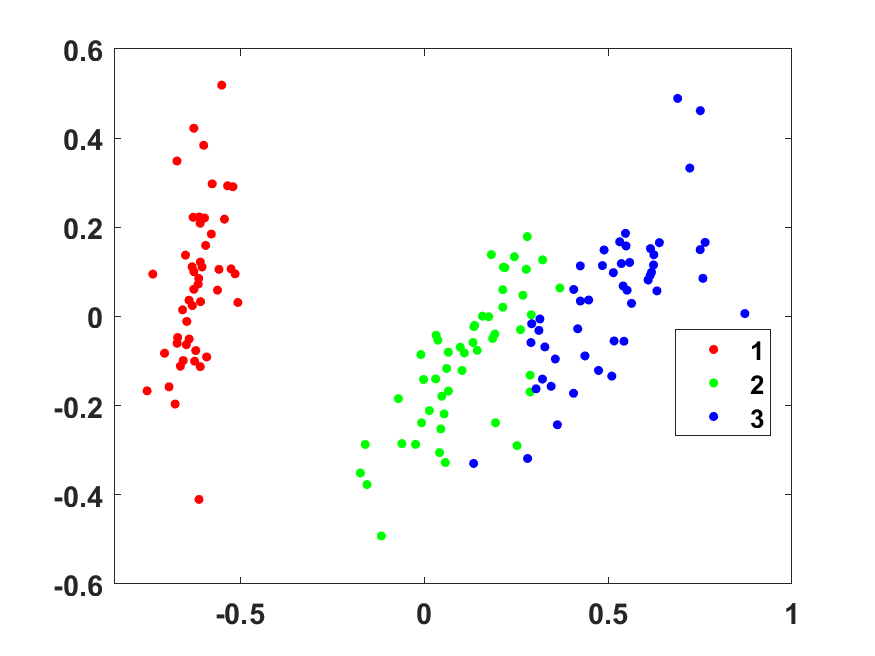}
    }
    \subcaptionbox{Noise Suppression: 93.55\% \label{Iris Noise SVM}}{
        \centering
        \includegraphics[scale=0.28]{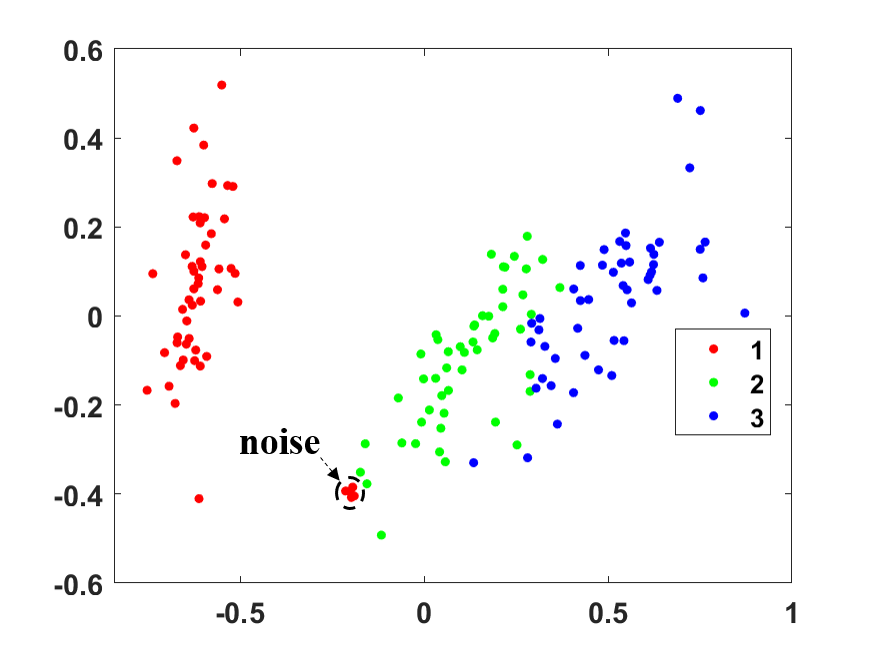}
    }
    \subcaptionbox{Rectified by $\pi$-noise: 93.75\% \label{Iris Pi-noise SVM}}{
        \centering
        \includegraphics[scale=0.28]{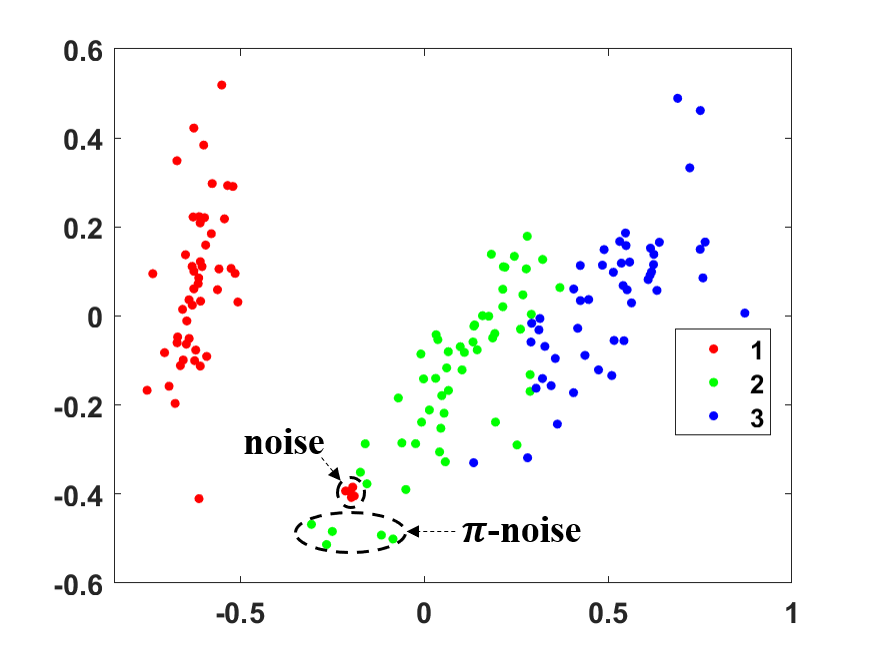}
    }
    \subcaptionbox{Excessive $\pi$-noise: 92.73\% \label{Iris over Pi-noise SVM}}{
        \centering
        \includegraphics[scale=0.28]{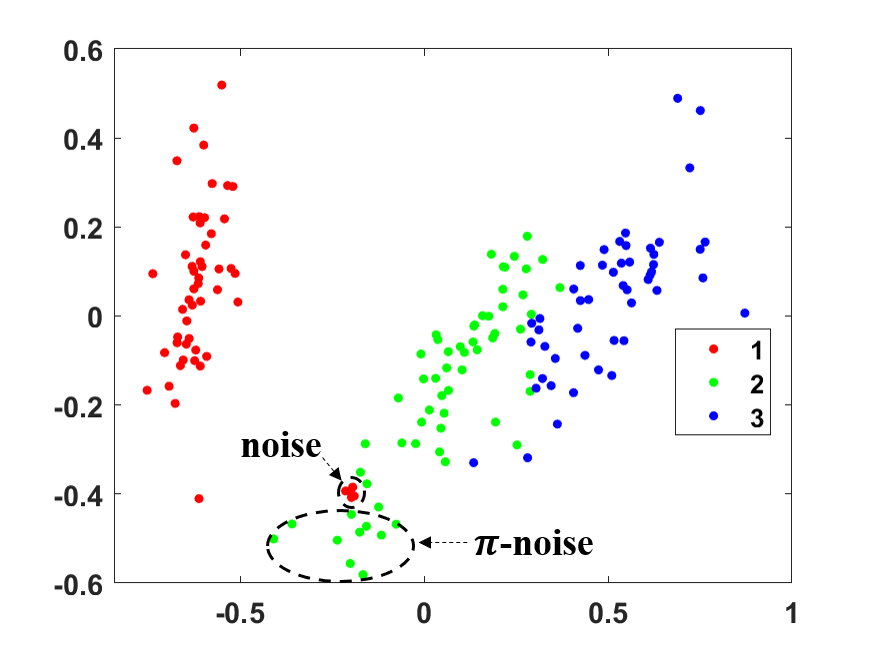}
    }
    
    \subcaptionbox{Wine ACC: 97.19\% \label{Wine SVM}}{
        \centering
        \includegraphics[scale=0.28]{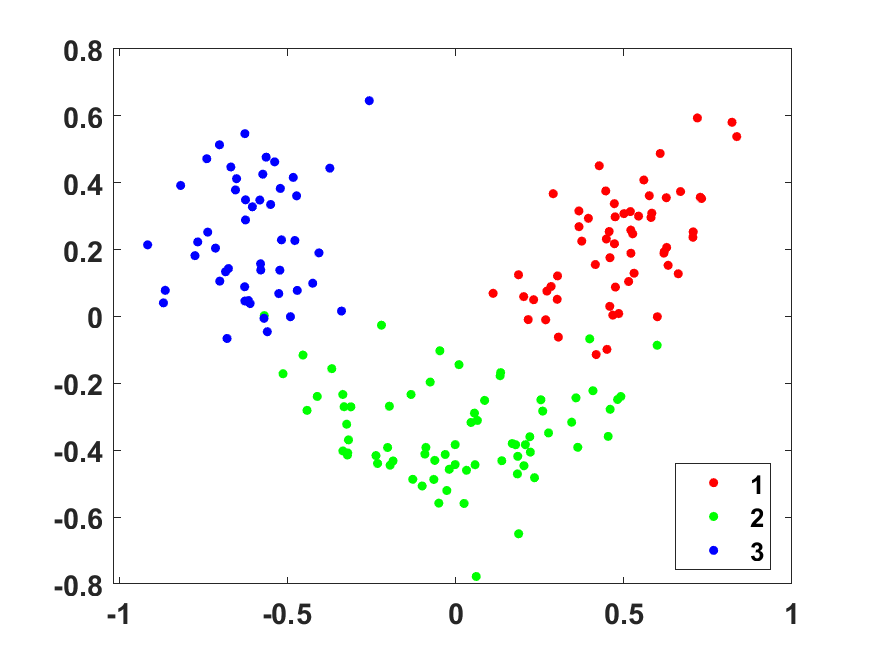}
    }
    \subcaptionbox{Noise Suppression: 94.54\% \label{Wine Noise SVM}}{
        \centering
        \includegraphics[scale=0.28]{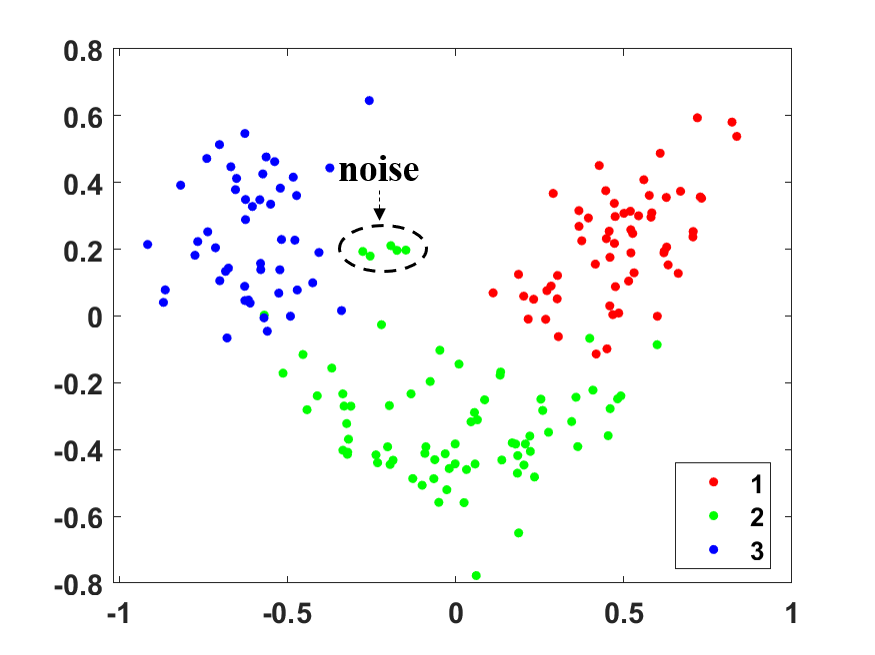}
    }
    \subcaptionbox{Rectified by $\pi$-noise: 95.68\% \label{Wine Pi-noise SVM}}{
        \centering
        \includegraphics[scale=0.28]{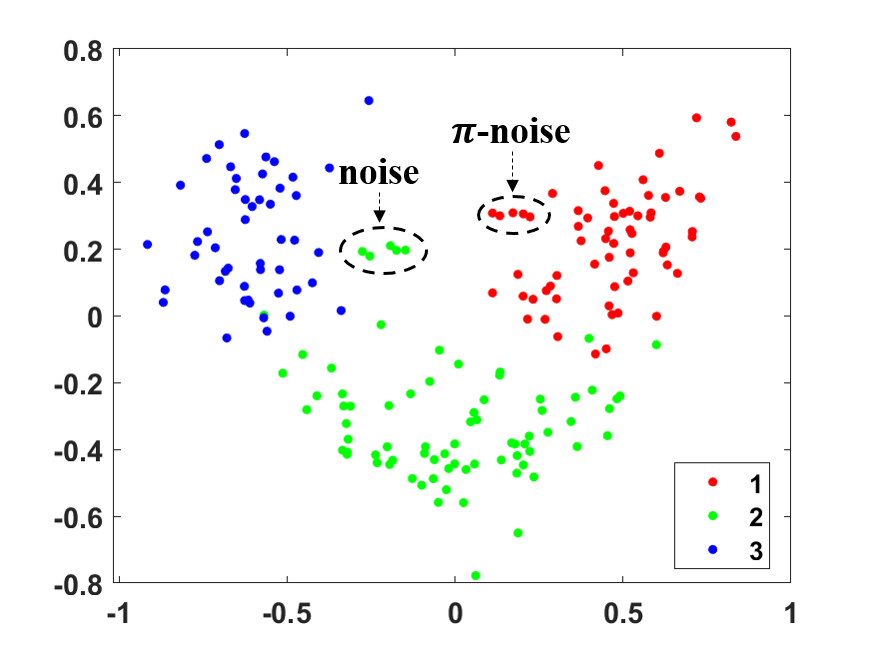}
    }
    \subcaptionbox{Excessive $\pi$-noise: 94.58\% \label{Wine over Pi-noise SVM}}{
        \centering
        \includegraphics[scale=0.28]{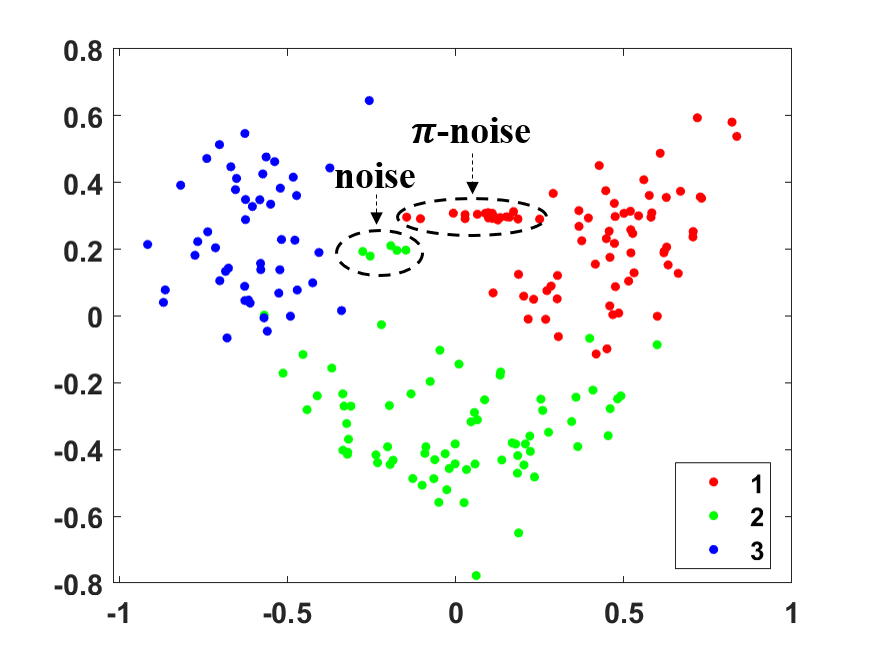}
    }
    
    
    \caption{The accuracy of SVM on benchmark datasets with the rectified $\pi$-noise. The rows suggest the classification results on Toy, Iris, and Wine datasets, respectively. The first column is the result of the original dataset. The second column is the result of the dataset with the Gaussian random noise. The third column shows the proper number of rectified $\pi$-noises introduced to rectify the performance. The last column suggests that too many rectified $\pi$-noises can also degrade the performance.}
    \label{SVM Classification Accuracy}
\end{figure*}

\begin{figure*}[t]
    \centering
    \subcaptionbox{Toy ACC: 100.00\% \label{Toy kmeans}}{
        \centering
        \includegraphics[scale=0.28]{./figure/toy/x.png}
    }
    \subcaptionbox{Noise Suppression: 90.91\% \label{Toy Noise kmeans}}{
        \centering
        \includegraphics[scale=0.28]{./figure/toy/supno_x1.png}
    }
    \subcaptionbox{Rectified by $\pi$-noise: 92.00\% \label{Toy Pi-noise kmeans}}{
        \centering
        \includegraphics[scale=0.28]{./figure/toy/suppino_x1.png}
    }
    \subcaptionbox{Excessive $\pi$-noise: 76.92\% \label{Toy over Pi-noise kmeans}}{
        \centering
        \includegraphics[scale=0.28]{./figure/toy/supoverpino_x1.png}
    }
    
    
    \subcaptionbox{Iris ACC: 88.67\% \label{Iris kmeans}}{
        \centering
        \includegraphics[scale=0.28]{./figure/iris/x.png}
    }
    \subcaptionbox{Noise Suppression: 88.39\% \label{Iris Noise kmeans}}{
        \centering
        \includegraphics[scale=0.28]{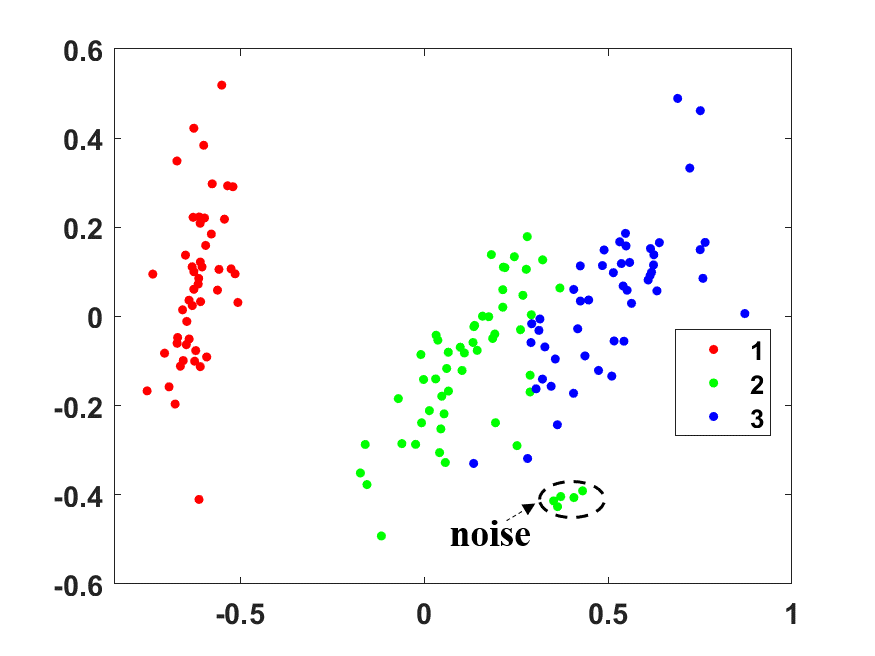}
    }
    \subcaptionbox{Rectified by $\pi$-noise: 89.38\% \label{Iris Pi-noise kmeans}}{
        \centering
        \includegraphics[scale=0.28]{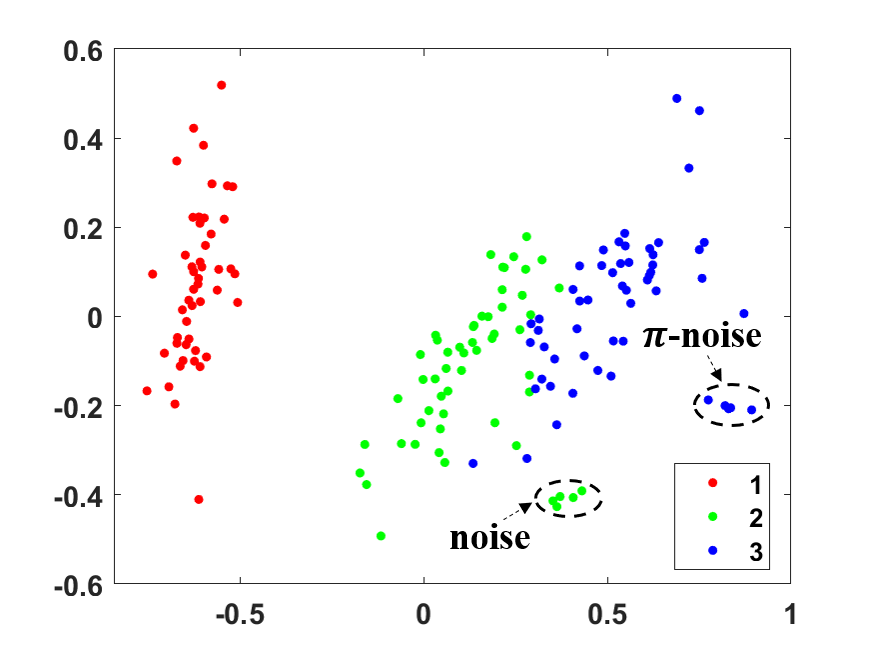}
    }
    \subcaptionbox{Excessive $\pi$-noise: 66.67\% \label{Iris over Pi-noise kmeans}}{
        \centering
        \includegraphics[scale=0.28]{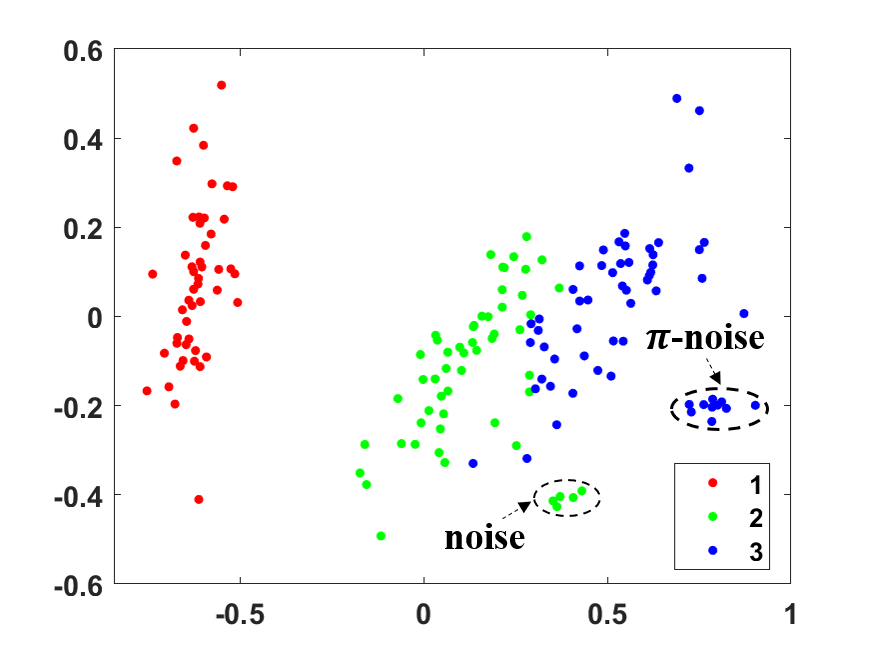}
    }
    
    \subcaptionbox{Wine ACC: 94.94\% \label{Wine kmeans}}{
        \centering
        \includegraphics[scale=0.28]{./figure/wine/x.png}
    }
    \subcaptionbox{Noise Suppression: 92.35\% \label{Wine Noise kmeans}}{
        \centering
        \includegraphics[scale=0.28]{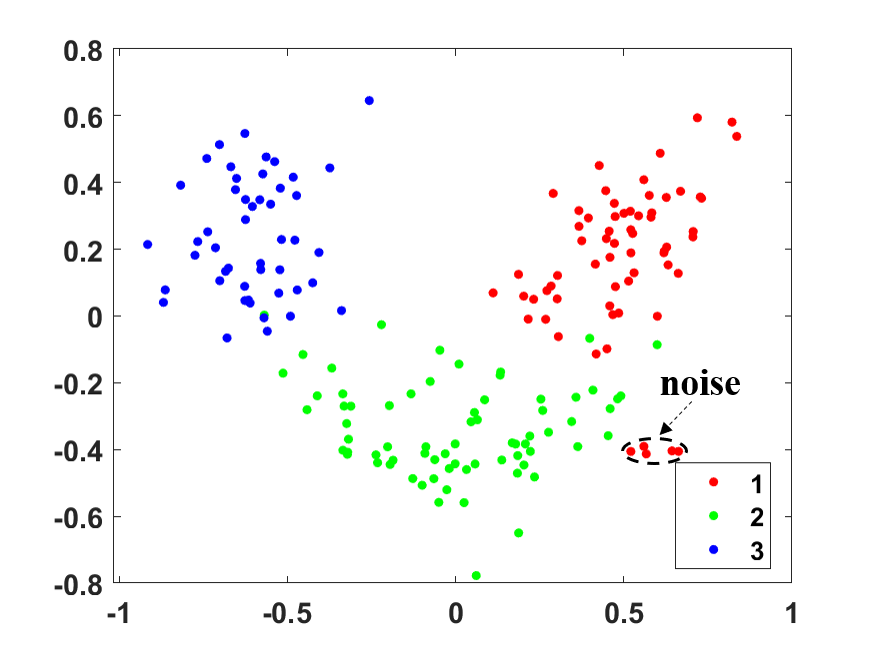}
    }
    \subcaptionbox{Rectified by $\pi$-noise: 94.58\% \label{Wine Pi-noise kmeans}}{
        \centering
        \includegraphics[scale=0.28]{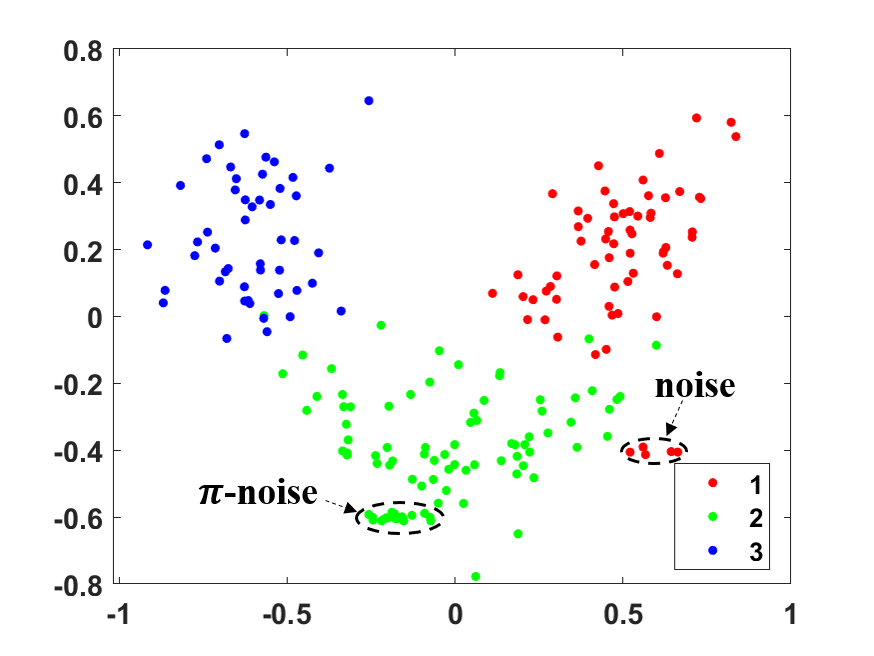}
    }
    \subcaptionbox{Excessive $\pi$-noise: 94.37\% \label{Wine over Pi-noise kmeans}}{
        \centering
        \includegraphics[scale=0.28]{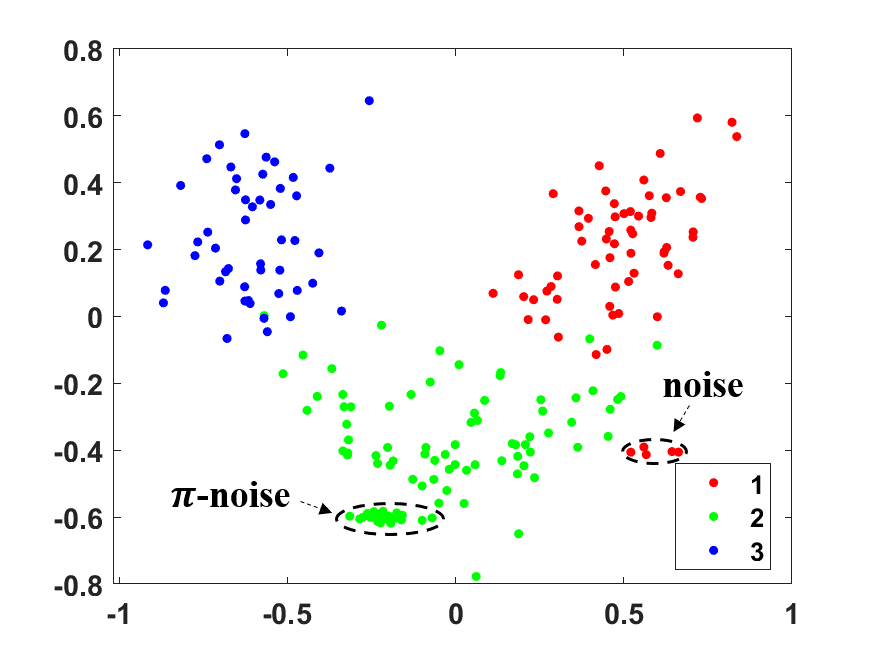}
    }
    
    
    \caption{The clustering accuracy of $K$-Means on benchmark datasets with the rectified noise. 
        The rows represent the clustering results on Toy, Iris, and Wine datasets, respectively. 
        The first column is the clustering result of the original dataset. 
        The second column is the result with Gaussian random noise. 
        The third column shows the proper number of rectified $\pi$-noises introduced to rectify the performance. 
        The last column indicates that too many rectified $\pi$-noises can also degrade the clustering.}
    \label{kmeans Clustering Accuracy}
    \vspace{-2mm}
\end{figure*}

\begin{itemize}
    \item \textit{Multiplicative Noise}: This noise generally is generated by the change of channel and can be represented as $u_\epsilon = u \times \epsilon$ where $u$ is the original signal. Meanwhile, the salt-and-pepper noise is common multiplicative noise for images. Specifically, due to the signal disturbed by the sudden strong interference or bit transmission error, the image generates unnatural changes such as the black pixels in bright areas or white pixels in dark areas. As shown in Fig. \ref{pepper noise 0}-\ref{pepper noise 0.5}, the training images are corrupted by the different degrees of salt-and-pepper noise. The degree means the proportion of pixels in the image is set to $0$ or $255$.
    \item \textit{Gaussian Noise}: This noise generally is added to the original signal such as $u_\epsilon = u + \epsilon$. However, different from the multiplicative noise, the additive noise is independent and stochastic. The Gaussian noise, one of the most common additive noises, obeys the Gaussian distribution $\mathcal{N}(\mu, \sigma^2)$. To introduce Gaussian noise reasonably, the image is normalized from $[0, 255]$ into $[0, 1]$. $\mu$ and $\sigma$ are chosen from $[0, 1]$. Then, the noise is sampled from the distribution and added to  the image. Finally, the generated images are recovered into $[0, 255]$. Gaussian noisy images are shown in Fig. \ref{Gaussian Original}-\ref{gaussian mean_0_5_sigma_0_5}.
    \item \textit{Uniform Noise}: This type of noise is generated from the uniform distribution ${U}(a, b)$ and added to the raw image. The same operation is adopted to add uniform noise. As shown in Fig. \ref{Uniform Original}-\ref{uniform low_0_high_1}, the noisy images are listed according to different $a$ and $b$.
    \item \textit{Dimension Noise}: This type of noise is obtained by random linear transformation and nonlinear activation of the original data, which is cascaded behind the original data subsequently. It can be written as $\bm u_\epsilon = \bm u \| \textrm{sgn}(\bm P \bm u)$, where $\|$ is the concat operation and $\bm P$ is a linear transformation.
\end{itemize}

In the experiment, the multiplicative noise with degree=3, additive noise with $\mu=0.5,\sigma=0.5$, and uniform noise with $a=0, b=1$ are chosen as the enhanced noise. For dimension noise, $\bm p$ is generated from a random uniform distribution and $sgn(\cdot)$ is a sign function.

\subsubsection{Experiment Results}
To show the effect of different noises on the model, 
the experiments with the different noisy proportion, $p$, from $\left\{ 0.0,0.05, \ldots, 0.95 \right\}$ 
are conducted. 
The results are shown in Fig. \ref{Classification Accuracy}.
From the three figures, a counterintuitive conclusion is obtained: 
Data with a little simple random noise enhance the model, compared with 
the ``noiseless'' data. The curve is an inverted U-shape curve, 
which indicates that proper noise is beneficial. 
From the visualization in Fig. \ref{Noise Image}, 
it is easy to find that the proper random noise blurs the background and remains 
the necessary feature of the airplane, leading to a decreasing complexity. 
Besides, the enhancement of dimension noise is shown in Table \ref{demension_results}. 
Among them, $m$ is the dimension of the added noise. 
The classification performance of original data is substantially improved by adding noise.
The experimental results verify the existence of $\pi$-noise and support the guess 
about the amount of $\pi$-noises.

\begin{table}[t]
    \centering
    \renewcommand\arraystretch{1.2}
    \caption{Datasets}
    \label{table_datasets}
    \begin{tabular}{lccc}
        \hline
        
        \hline
        \textbf{Datasets} & \textbf{\# Samples} & \textbf{\# Features} & \textbf{\# Classes} \\ 
        \hline
        Toy               & 200                    & 2                 & 2                \\
        IRIS              & 150                    & 2                 & 3               \\
        Wine              & 178                    & 13                & 3                \\
        \hline
        
        \hline
    \end{tabular}
\end{table}

\subsection{Rectified $\pi$-Noise}
Another application is to use the $\pi$-noise to neutralize the negative effect 
of the pure noise. 
Instead of detect and eliminate the pure noise in data points, 
another scheme is to add some $\pi$-noises to rectify the data distribution. 
It is particularly preferable for incremental learning systems. 
The core assumption of incremental systems is the expensive re-training. 
When a batch of data points with some noisy points come, the system suffers 
from the irreversible damage and the idea to add $\pi$-noise provides a 
cheap scheme. 
It corresponds to the gum-nail instance proposed in Section \ref{section_introduction}. 
Before the details of experiments, it should be emphasized that 
the noise added in this subsection is actually \textit{noisy instances}, rather than 
the additive or multiplicative noise acting on the original data instances.

\subsubsection{Datasets Setting}
The experiments are conducted on three tasks, including classification, 
clustering, dimensionality reduction. 
For classification and clustering, totally 3 datasets are utilized to investigate the performance of rectified $\pi$-noise, 
including a synthetic dataset and 2 UCI \cite{UCI} datasets. 
For each dataset, each class has the same number of samples. 
The details of these datasets are reported in Table \ref{table_datasets}. 
The toy dataset, namely Toy, is sampled from two Gaussian distributions, 
$\mathcal{N}([0.3;0.3],[0.01, 0; 0, 0.01])$ and $\mathcal{N}([0.7;0.7],[0.01, 0; 0, 0.01])$. 
Each class consists of $100$ samples. 
To show the result more vividly, all datasets are projected into two-dimensional spaces with the principal component analysis (PCA) \cite{PCA}.
For dimensionality reduction, 
the experiments are conducted on a toy dataset, which contains two classes 
in two-dimensional space. 
Among them, class 1 has $100$ data points sampled from $\mathcal{N}([0.3; 0.3], [0.01, 0; 0, 0.01])$ 
and class 1 has the same number of data points sampled from $\mathcal{N}([1.3; 0.3], [0.01, 0; 0, 0.01])$.

\subsubsection{Classification (Support Vector Machine {\rm \cite{SVM}})} 
In the experiments, the classical support vector machine (SVM) is utilized as the classifier. 
Meanwhile, the One-versus-Rest strategy is equipped for SVM to handle the dataset with multiple classes. 
Firstly, SVM is run on the original benchmark dataset. 
Secondly, to show the degradation of classification accuracy on the noisy dataset, 
the datasets are equipped with the noisy samples generated from the different Gaussian distributions. 
For Toy dataset, $20$ noisy points are sampled from $\mathcal{N}([0.5;0.8], [0.001, 0; 0, 0.001])$. 
Each dataset in UCI datasets are added $5$ noises. 
Among them, 
the noises in Iris are sampled from $\mathcal{N}([-0.2;-0.4], [0.01, 0; 0, 0.01])$, 
and the noises in Wine are sampled from $\mathcal{N}([-0.2;0.2], [0.05, 0; 0, 0.01])$. 
Thirdly, the rectified $\pi$-noise is introduced and SVM predicts the classification to verify the rectified capability. 
Among them, Toy dataset is equipped with $20$ synthetic samples drawn from $\mathcal{N}([0.8;0.2], [0.001, 0; 0, 0.001])$, 
Iris dataset is equipped with $5$ data points sampled from $\mathcal{N}([-0.2;-0.5], [0.10, 0; 0, 0.05])$, 
and Wine dataset is equipped with $5$ samples drawn from $\mathcal{N}([0.1,0.3], [0.10, 0; 0, 0.01])$. 
Lastly, the number of $\pi$-noise is increased to explore the impact on performance from the number.
The results are shown in Fig. \ref{SVM Classification Accuracy}.

\subsubsection{Clustering ($K$-Means {\rm \cite{KMeans}})}
The noisy sample set consists of $20$ samples from class 1 and 
are generated from $\mathcal{N}([1.3;1.0], [0.001, 0; 0, 0.001])$. 
More importantly, the rectified $\pi$-noises consist of $20$ points from class 2 and 
are drawn from $\mathcal{N}([0.3;1.0], [0.001, 0; 0, 0.001])$.
The results are shown in Fig. \ref{kmeans Clustering Accuracy}.

\subsubsection{Dimensionality Reduction (Linear Discriminant Analysis {\rm \cite{LDA}})}
Finally, the classical linear discriminant analysis (\textit{LDA}) \cite{LDA} is employed to find 
whether the $\pi$-noise could rectify the performance of dimensionality reduction. 
Firstly, the toy dataset is produced which contains two classes in two-dimensional space. 
Among them, class 1 has $100$ points sampled from $\mathcal{N}([0.3; 0.3], [0.01, 0; 0, 0.01])$, 
and class 1 has the same number of points sampled from $\mathcal{N}([1.3; 0.3], [0.01, 0; 0, 0.01])$. 
Meanwhile, the noises have $20$ points of class 1 and are generated from $\mathcal{N}([1.3;1.0], [0.001, 0; 0, 0.001])$. 
More importantly, the rectified $\pi$-noises are composed of $20$ samples from 
class 2 and are sampled from $\mathcal{N}([0.3;1.0], [0.001, 0; 0, 0.001])$.
The results are shown in Fig. \ref{LDA}.

\subsubsection{Experimental Results} 
From three types of learning models, it is easy to conclude that 
there exists $\pi$-noise eliminate the negative effect of pure noise 
and rectifying the learning systems. 
It enlightens us that adding some proper random noisy points, 
instead of detecting the existing pure noise and removing it, 
may also help to improve the performance, which offers a new scheme for investigation of robust models.


\section{Future Works} \label{section_future_work}
The discussions in this paper are elementary and instructive. 
More detailed analysis and investigations deserve further attention in the future. 
For instance, there are several attractive topics listed as follows:

\begin{itemize}
    \item Although the $\pi$-noise widely exists in different fields, 
		there is a crucial question: 
		\textbf{What property will the ($\alpha$-strong) $\pi$-noise have?} 
        For example, it is promising to study which kind of random noise 
        (\textit{e.g.}, uniform noise, Gaussian noise) 
        is more likely to be $\pi$-noise in diverse scenes. 
        It will be a core in the future investigations. 
    \item As highlighted in the preceding sections, although a little $\pi$-noise 
        enhances the performance, too much $\pi$-noise would 
        lead to degeneration as well. What is the relationship between 
        the quantity of $\pi$-noise and inflection point of performance?
		In other words, \textbf{what is the upper bound of the quantity 
        of $\pi$-noise that maximizes ${\rm MI}(\mathcal{T}, \bm \epsilon)$}? 
        For multivariate Gaussian noise, the problem is equivalent to 
        find a rigorous upper-bound of the covariance matrix regarding certain norm. 

        \begin{figure*}[t]
            \centering
            \subcaptionbox{Original \label{Original_LDA}}{
                \centering
                \includegraphics[scale=0.35]{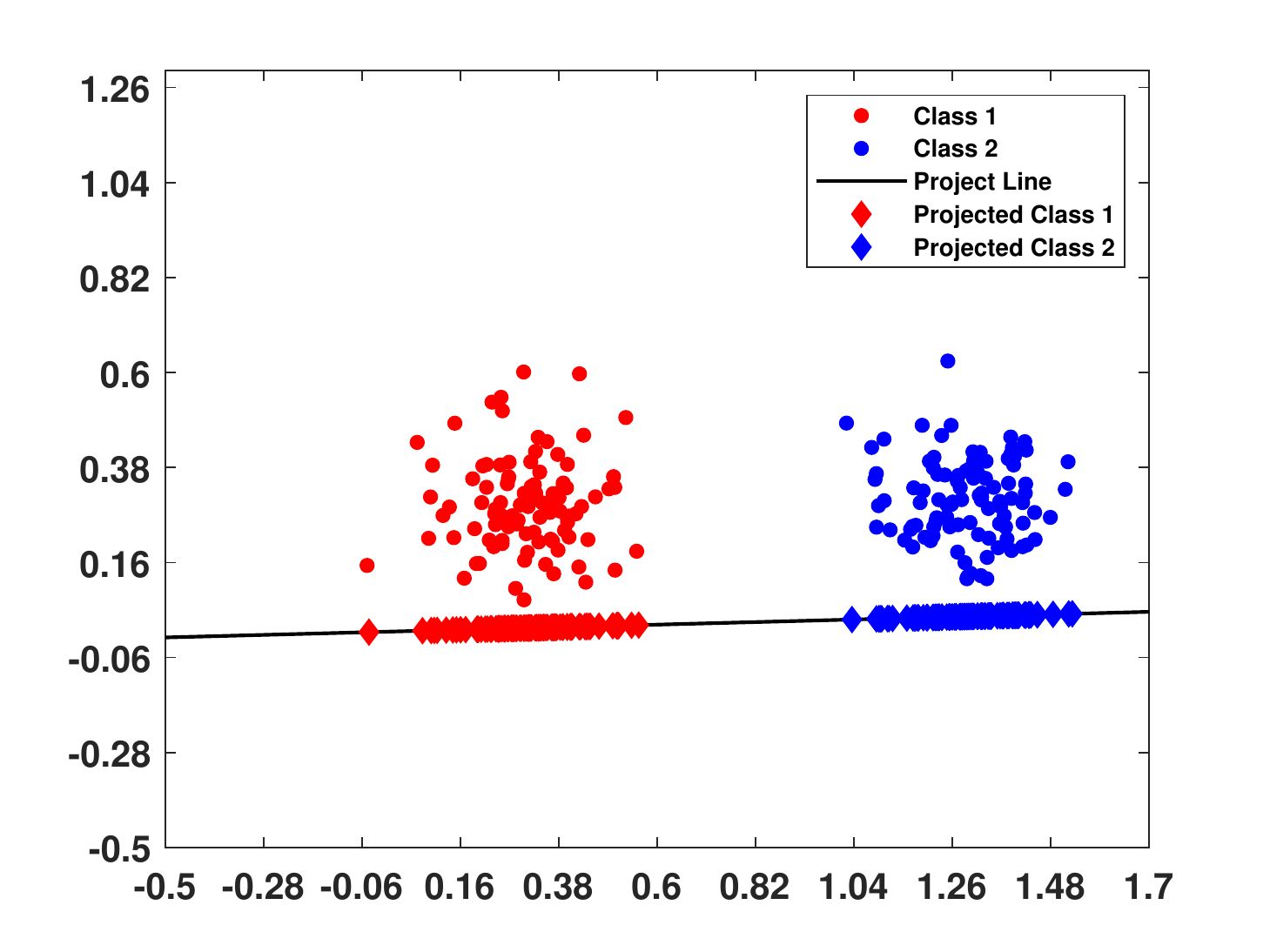}
            }
            \subcaptionbox{Polluted by Noise \label{Noise_LDA}}{
                \centering
                \includegraphics[scale=0.35]{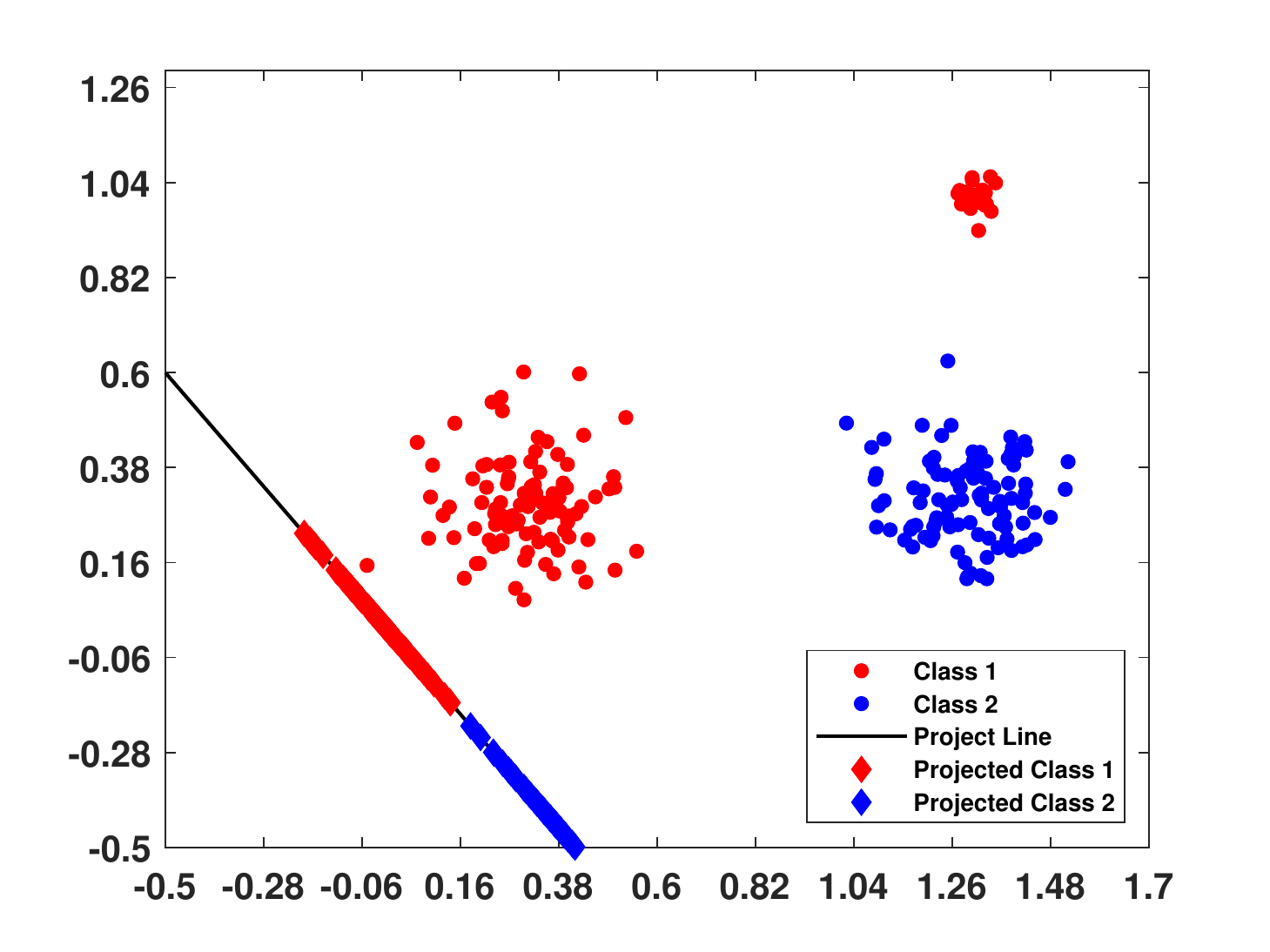}
            }
            \subcaptionbox{Rectified by $\pi$-Noise \label{PINoise_LDA}}{
                \centering
                \includegraphics[scale=0.35]{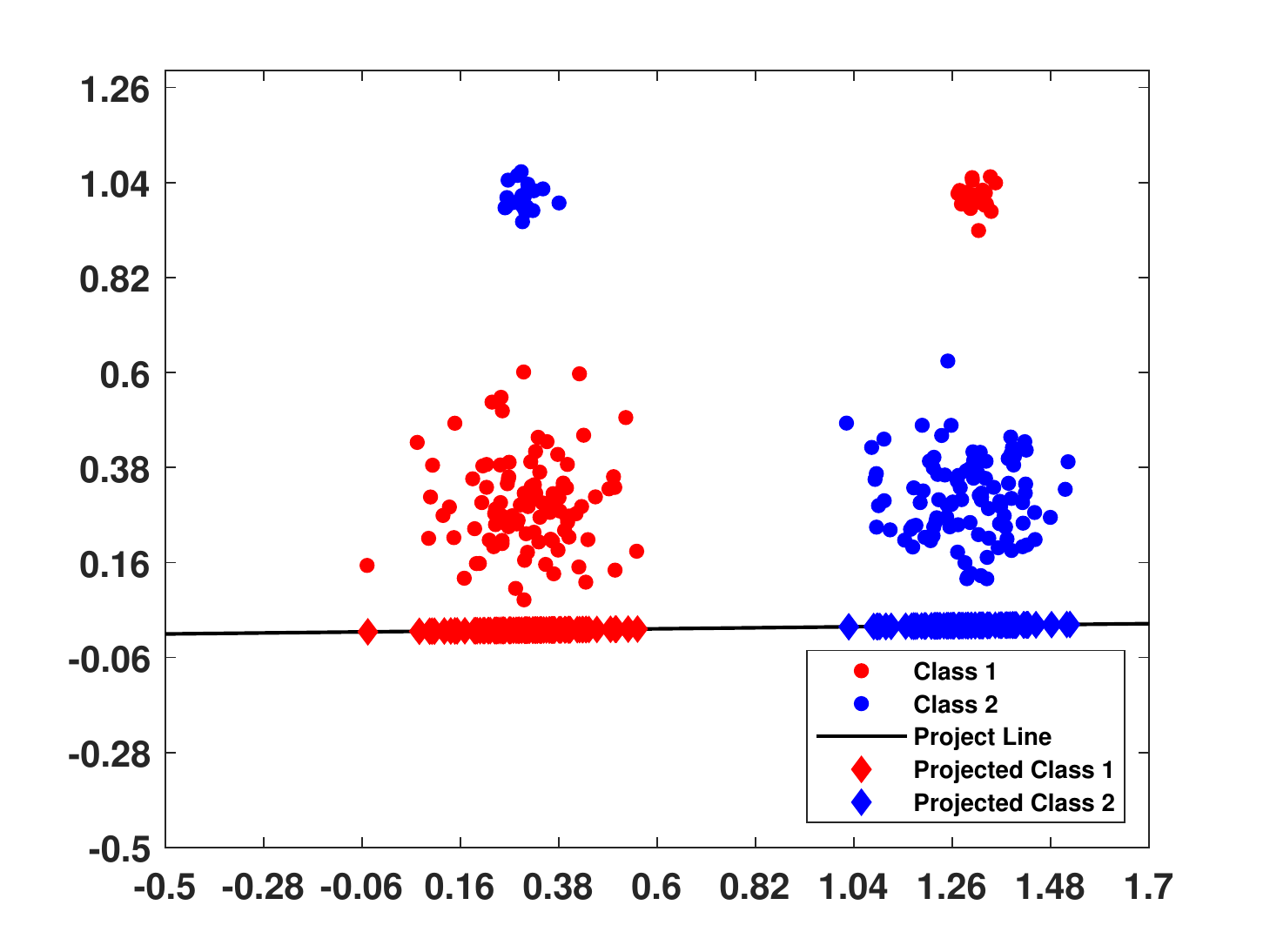}
            }
            \caption{LDA on the toy dataset. Fig. \subref{Original_LDA} 
                visualizes the result of LDA on original data. 
                Fig. \subref{Noise_LDA} shows that the noises disturb the results and 
                Fig. \subref{PINoise_LDA} indicates that the $\pi$-noise can rectify the 
                projection direction.}
            \label{LDA}
            \vspace{-2mm}
        \end{figure*}

	\item Although the existence of $\pi$-noise has been verified in some cases (\textit{e.g.}, classification, stochastic resonance), 
		\textbf{how to prove the existence of $\pi$-noise under general settings} is still an attractive problem. 
    \item As shown in Section \ref{section_pi_noise_classification}, the computation of task entropy 
        offers a new way to measure the complexity of datasets. 
        Therefore, it is attractive to study whether the measurement induced by $\pi$-noise could provide 
        a novel and practical framework of learning theory like the Rademacher complexity \cite{UnderstandingML}. 
        It may also show how to measure the ability to provide information per unit data size, namely information capacity. 
    \item Although the rectification ability of $\pi$-noise is sufficiently shown, 
        how to find the rectified $\pi$-noise is an urgent problem. 
        One way that may work is to find the desirable distribution via variational methods. 
	\item $\pi$-noise could be a new principle for designing models. 
        For instance, adversarial training can be more efficient if 
        the optimization of $\bm \epsilon$ aims at finding $\pi$-noise. 
        A simple loss incorporating $\pi$-noise is 
        \begin{equation}
            \min_{\theta} \max _{\bm \epsilon} \sum_{\bm x \in \bm X} \ell(f_{\bm \theta}(\bm x + \bm \epsilon), \bm y) +  {\rm MI}(\mathcal{T}, \bm \epsilon) . 
        \end{equation}
        Compared with the heuristic search of $\bm \epsilon$, the above principle 
        may be more reliable and stable. 
        In object detection, $\pi$-noise could provide a reliable principle to 
        expand the bounding box to promote the detection by incorporating positive background information. 
    \item The clear difference between $\pi$-noise and pure 
        noise also inspires us to rethink the data preprocessing. 
        The existence of $\pi$-noise and its definition based on tasks imply that 
        the denoising scheme should be designed for specific tasks since 
        some noises may be beneficial.  
    \item $\pi$-noise will be the core of Vicinagearth Security \cite{VS}. For example, in the field of non-line-of-sight imaging and underwater imaging, 
        the theory of $\pi$-noise may provide a new perspective 
        to view received signals and help to design a stronger imaging system. 
        $\pi$-noise also plays an important role in UAV (unmanned aerial vehicle) applications. 
        How to apply the theory of $\pi$-noise to Vicinagearth Security will be an emphasis of future works. 
\end{itemize}

In sum, it requires more systematic and rigorous investigations of $\pi$-noise 
in the future.

\section{Conclusion} 
This paper rethinks whether the noise always results in a negative impact. 
The doubt comes from the loose definition of noise. 
Through modeling the mutual information of task $\mathcal{T}$ and noise $\bm \epsilon$, 
the traditional ``noise'' can be classified into two categories, 
$\pi$-noise and pure noise. 
In brief, $\pi$-noise is the random signal that can simplify the target task. 
By conducting some convincing experiments and showing that 
some existing topics (\textit{e.g.}, stochastic resonance, multi-task learning, adversarial training) 
can be explained as special cases, 
we empirically and theoretically conclude that $\pi$-noise is ubiquitous in diverse fields. 
There are still plenty of attractive problems that deserves more investigations, 
including but not limited to the general property of $\pi$-noise, 
the upper-bound of quantity of $\pi$-noise, 
the existence of $\pi$-noise under general settings, 
the new principle for designing models regarding $\pi$-noise, 
\textit{etc}.
Importantly, $\pi$-noise is also related to the study of information capacity. 
Both of them will be theoretical bases of Vicinagearth Security, 
which is the core of my future works.

\bibliographystyle{IEEEtran}

\bibliography{citations.bib}

\begin{IEEEbiography}[{\includegraphics[width=1in,height=1.25in,clip,keepaspectratio]{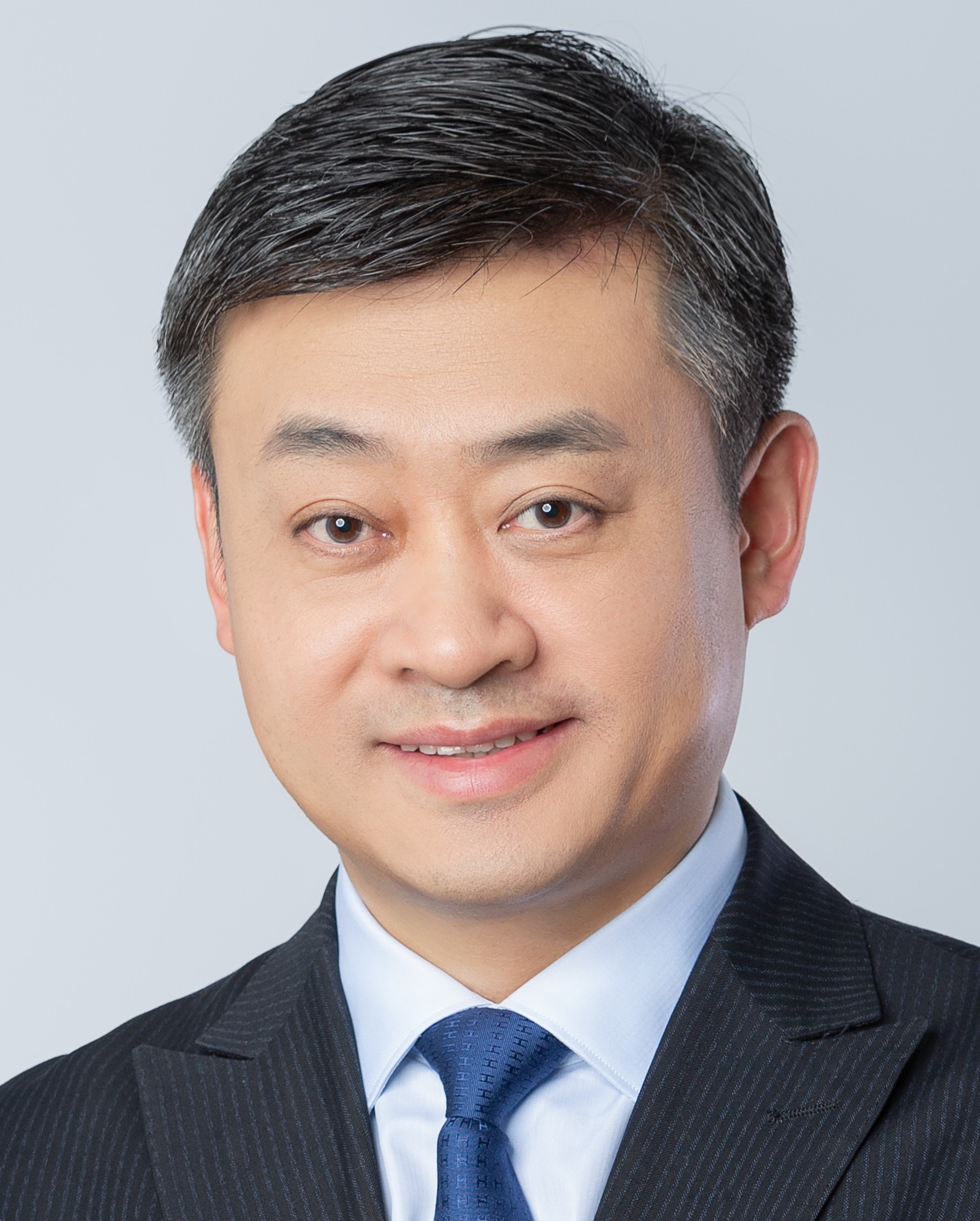}}]{Xuelong Li} (M'02-SM'07-F'12)
    is a full professor with the School of Artificial Intelligence, OPtics and ElectroNics (iOPEN), 
    Northwestern Polytechnical University, Xi'an, China. 
\end{IEEEbiography}   

\end{document}